\documentclass[journal, twoside]{IEEEtran}
\let\labelindent\relax
\usepackage{graphicx}
\usepackage{subcaption}
\usepackage{eqlist}
\usepackage{amsfonts}
\usepackage{tabularx}
\usepackage{booktabs}
\usepackage{amssymb}
\usepackage{amsmath}
\usepackage{enumitem}
\usepackage{mathtools}
\usepackage{threeparttable}
\usepackage[lined, ruled, linesnumbered, commentsnumbered]{algorithm2e}
\usepackage{lipsum}

\usepackage[usenames,dvipsnames]{xcolor}
\usepackage{comment}

\usepackage{multirow}

\usepackage{CJKutf8}

\begin{document}
\title{A Dual-arm Robot that Autonomously Lifts Up and Tumbles Heavy Plates Using Crane Pulley Blocks}
\author{Shogo Hayakawa$^1$, Weiwei Wan$^{1*}$, Keisuke Koyama$^1$ and Kensuke Harada$^{1,2}$
\thanks{$^{1}$Graduate School of Engineering Science, Osaka University, Japan.}%
\thanks{$^{2}$National Inst. of AIST, Japan.}%
\thanks{Contact: Weiwei Wan, {\tt\small wan@hlab.sys.es.osaka-u.ac.jp}}
}

\markboth{Preprints at Arxiv, Submitted to a journal for review, 2020}
{Hayakawa \MakeLowercase{\textit{et al.}}: A Dual-arm Robot that Autonomously Lifts Up and Tumbles Heavy Plates Using Crane Pulley Blocks} 
\maketitle

%%%%%%%%%%%%%%%%%%%%%%%%%%%%%%%%%%%%%%%%%%%%%%%%%%%%%%%%%%%%%%%%%%%%%%%%%%%%%%%%
\begin{abstract}
This paper develops a planner that plans the action sequences and motion for a dual-arm robot to lift up and flip heavy plates using crane pulley blocks. The problem is motivated by the low payload of modern collaborative robots. Instead of directly manipulating heavy plates that collaborative robots cannot afford, the paper develops a planner for collaborative robots to operate crane pulley blocks. The planner assumes a target plate is pre-attached to the crane hook. It optimizes dual-arm action sequences and plans the robot's dual-arm motion that pulls the rope of the crane pulley blocks to lift up the plate. The crane pulley blocks reduce the payload that each robotic arm needs to bear. When the plate is lifted up to a satisfying pose, the planner plans a pushing motion for one of the robot arms to tumble over the plate while considering force and moment constraints. The article presents the technical details of the planner and several experiments and analysis carried out using a dual-arm robot made by two Universal Robots UR3 arms. The influence of various parameters and optimization goals are investigated and compared in depth. The results show that the proposed planner is flexible and efficient.
\end{abstract}


%%%%%%%%%%%%%%%%%%%%%%%%%%%%%%%%%%%%%%%%%%%%%%%%%%%%%%%%%%%%%%%%%%%%%%%%%%%%%%%%
\section{Introduction}
\IEEEPARstart{A}{lthough} modern industrial robots can lift as much as 1000$kg$ payload. They tend to be expensive, bulky, dangerous, and have to be fenced in a work cell to keep human workers safe. They are not suitable for human-intensive manufacturing sites. On the other hand, collaborative robots are developed to work together with humans, they have a smaller size and are considered safe. However, they tend to have small payloads to ensure a good response to collisions. The heavy load and collaboration form a trade-off that leads to an interesting and unsolved problem -- how to use collaborative robots to manipulate heavy objects.

Previously, researchers in robot manipulation suggested working on heavy objects using non-prehensile manipulation, which leverages external supports to afford a part of the total workload \cite{yoshida2005}\cite{murooka2015}\cite{ohashi2016realization}. Despite the cleverness, the masses of the target objects to be manipulated remain limited. They must meet the force requirements from external supports. Also, carrying out complicated manipulation tasks like turning over the object is difficult and deserves sophisticated optimization, and planning \cite{hou2018}\cite{haustein2019placing}. Under this background, we explore new manipulation methods to work on heavy objects using collaborative robots. 

\begin{figure}
    \centering
    \includegraphics[width=.95\linewidth]{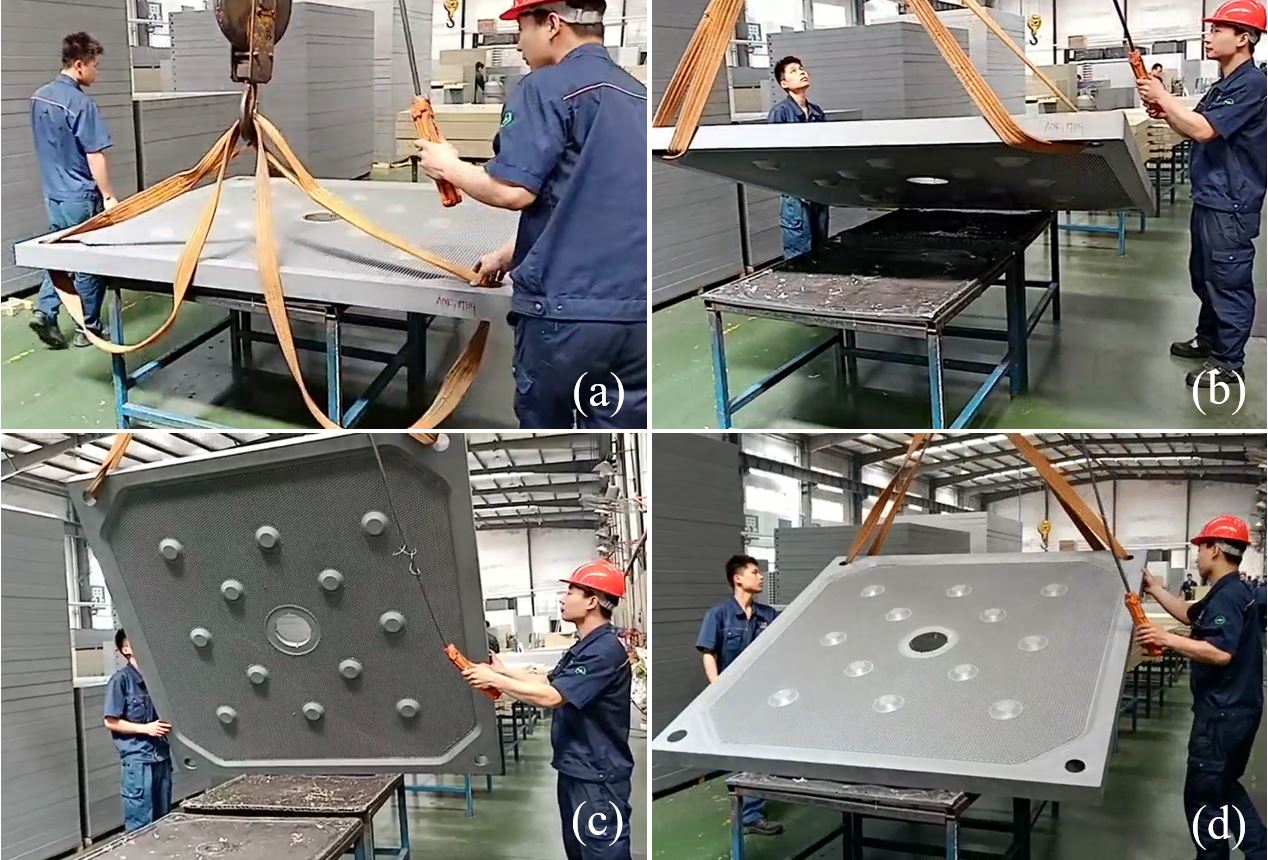}
    \caption{Human workers in a sewage press machine factory flip press boards with the help of a gantry crane. (a) Attaching the board to the crane hook using bearing belts. (b) Activating the crane to lift up the board. (c) The board is lifted up to a satisfying pose. (d) The board is flipped and is returned to the workbench.}
    \label{figmotivation}
\end{figure}

The conception we have in mind is to take robots as humans and plan them to manipulate various objects using humans' tools. Especially for heavy objects, we propose to use crane pulley blocks as the tool. The policy is inspired by a production process seen in a factory that produces sewage press machines. Sewage press machines are used to dehydrating coals. The pressboard of sewage press machines could be as heavy as 1000$kg$. In a factory that produces sewage press machines, as is shown in Fig.\ref{figmotivation}, human workers need to flip and clean both sides of the board before installing them to the cylinder axis. The factory developed a production procedure where human workers perform the flipping task using a gantry crane. They attach the board to the crane hook using bearing belts, activate the crane to lift up the board. When the board is raised to a satisfying pose, as is shown in Fig.\ref{figmotivation}(c), the human workers turn the board over by pushing it. Motivated by human workers' actions in the sewage machine factory, we in this research expect to perform a similar task using collaborative robots that operate crane pulley blocks. 

We develop a planner that optimizes the action sequences and plans the motion for a collaborative dual-arm robot to operate crane pulley blocks and lift up and tumble heavy plate-shaped objects with the crane pulley blocks' help. We assume target plates are pre-attached to the crane cook. Depth vision is used to recognize the position and orientation of the crane rope and the plates. Based on the recognized rope and plate poses, we optimize the initial and goal pulling poses for the rope and plan the robot's dual-arm motion that pulls the crane rope to lift up the plates. When the plates are lifted up to a satisfying pose, we compute the relations between contact wrenches and gravitational forces, and plan a pushing motion for one of the robot arms to tumble over the plates. 

Our main contribution is twofold. First, we initially develop a dual-arm robot that manipulates a plate much heavier than the maximum end-effector load by operating crane pulley blocks. Second, we formulated the constraints in using the crane pulley blocks and performed multiple optimizations to let the robot autonomously determine the pulling arms, pulling actions, and the tumbling trajectories. We present the technical details of the planner and several experiments and analysis carried out using the dual-arm robot and different plates. The influence of various parameters and optimization goals are investigated and compared in depth. The results show that the proposed planner is flexible and efficient.

In the remaining part of this article, we first review related work in Section II. Then, we present an overview of the method in Section III. After that, we show the technical details in Section IV, with experiments and analysis carried out using a dual-arm robot made by two Universal Robots UR3 arms and plates with different parameters in Section V. Conclusions and future work are presented in the last section.

%%%%%%%%%%%%%%%%%%%%%%%%%%%%%%%%%%%%%%%%%%%%%%%%%%%%%%%%%%%%%%%%%%%%%%%%%%%%
\section{Related Work}
\label{section2}

We review the related work in the manipulation planning of heavy objects. They are divided into two categories. The first category concentrates on firmly grasped or firmly attached objects. The goal is to keep balance or avoid breaking weak joints. For example, Harada et al. \cite{harada2005} used the Zero Moment Point (ZMP) of the humanoid robot-object system to compensate a humanoid robot's motion, and thus enable the humanoid robot to lift up an carry a heavy object stably. Urakubo et al. \cite{urakubo2014} studied lifting up a heavy object with small joint torques. They take advantage of singular configurations to avoid breaking the joints of a two-link robot. Berenson et al. \cite{berenson2009manipulation} formulated the manipulation planning of heavy objects as a probabilistic sampling problem while considering constrained manifolds. Yang et al. \cite{yang2018hardware} developed a tele-operating system considering mixed force and motion commands to avoid unexpected excessive force caused by massive payload. Zhu et al. \cite{zhu2020tmech} presented a planner for a wall climbing manipulator. The influence of self-weight was considered in determining the footholds for the suction modules. These studies assume that a robot firmly grasps the heavy object, which could be undesirable as heavy objects are generally large. A robotic gripper may not be able to grasp them with force or form closure. Also, even if a robot can grasp a part of a large, heavy object, it may remain challenging for the robot to pick up the object due to the fragile fingertip or fingerpad contacts. For this reason, non-prehensile manipulation is widely used to plan to manipulate heavy objects. It is the second category of studies we would like to review. 

Non-prehensile manipulation means manipulating an object without firmly grasping it. The representative non-prehensile manipulation primitives include pushing, sliding, tumbling, pivoting, lifting, caging, etc. Pushing was extensively studied by Mason \cite{mason1986}, and is currently widely used to manipulate the object. When a robot pushes an object, one needs to consider both the trajectories of the object and robot itself. Lynch and Mason \cite{lynch1996}, and more recently Zhou and Mason \cite{zhou2019pushing} analyzed this control and planning problem and implemented stable pushing. Recent progress in pushing include Zhou et al. \cite{zhou2017} and Yu et al. \cite{yu2016more}, which use probabilistic models to infer object states. Song et al. \cite{song2019object} proposed a nested approach to manipulate multiple objects together using pushing and learning. With the help of pushing, a robot can manipulate objects that cannot be directly grasped and lifted. For example, Murooka et al. \cite{murooka2015} presented a humanoid robot that pushed heavy objects by using whole-body contacts. Topping et al. \cite{topping2017quasi} modeled and planned a small quadruped robot's motion to open large doors.

Sliding is similar to pushing, but instead of exerting force sideways, sliding assumes pressing against the frictional surface. It also enables a small robot to manipulate large and heavy objects. Zhang et al. \cite{zhang2020} presented a dynamic model to plan the motion for a legged robot perform various sliding tasks like driving, inch worming, scooting, etc. Hang et al. \cite{hang2019pre} developed a pregrasp policy than slides thin objects to the corner of a table for easier pick-up. 

Tumbling means rotating an object while pressing it against a surface. Bai et al.\cite{bai2014dexterous} analyzed tumbling an object using a multi-fingered robotic hand. The motion is induced by a tilted palm and gravity. Fingers were used to protect the tumbling from overshooting. Cabezas et al. \cite{cabezas2020} presented a tumbling planner that accepts a given trajectory of rotation and computes the quasi-dynamic contacts. Correa et al. \cite{correa2019robust} and Cheng et al. \cite{cheng2019manipulation} respectively developed new usage of suction cups by considering using them as a tip for tumbling.

Rolling is a variation of tumbling, where the manipulated object is rotated continuously along a surface \cite{golubev2016insectomorphic}\cite{specian2018robotic}. 

Pivoting is a method that moves an object by leaving the object alternatively supported by corner points as if the object is walking on them. It is an extended version of tumbling. Aiyama et al. \cite{aiyama1993} seminally suggested the idea of robotic pivoting. Yoshida et al.\cite{yoshida2005} used a humanoid robot to pivot and move a heavy box.

Besides the primitives, researchers also employed high-level graph search to plan composite non-prehensile manipulation. For example, Maeda et al.\cite{maeda2005} proposed using a manipulation-feasibility graph to host the contact states \cite{xiao2001automatic} of an object and plan multi-finger graspless manipulation by searching the graph. Lee et al. \cite{lee2015hierarchical} proposed a hierarchical method that used a contact-state graph in lower layers to identify object-environment contacts, and determined robot contact sequences and maneuverings in a higher layer.

Compared with the above non-prehensile manipulation, our difference is as follows. Firstly, we do not stick to a robot itself. Our non-prehensile manipulation is performed by using crane pulley blocks to increase the duty of robots. Similar ideas could be found in a most up-to-date publication that use a collaborative robot to operate a manual pallet jack \cite{balatti2020}. We plan both the dual-arm robot action sequences and motion details to pull and return the crane pulley blocks' rope and ease non-prehensile flipping. Second, we plan a robot pushing motion to turn over heavy plates. We study a quasi-static prediction of a single contact, and use it to generate the trajectory to tumble plates.

%%%%%%%%%%%%%%%%%%%%%%%%%%%%%%%%%%%%%%%%%%%%%%%%%%%%%%%%%%%%%%%%%
\section{Overview of the Proposed Method}
\label{section3}

\begin{figure}[!htbp]
    \centerline{\includegraphics[width=.95\linewidth]{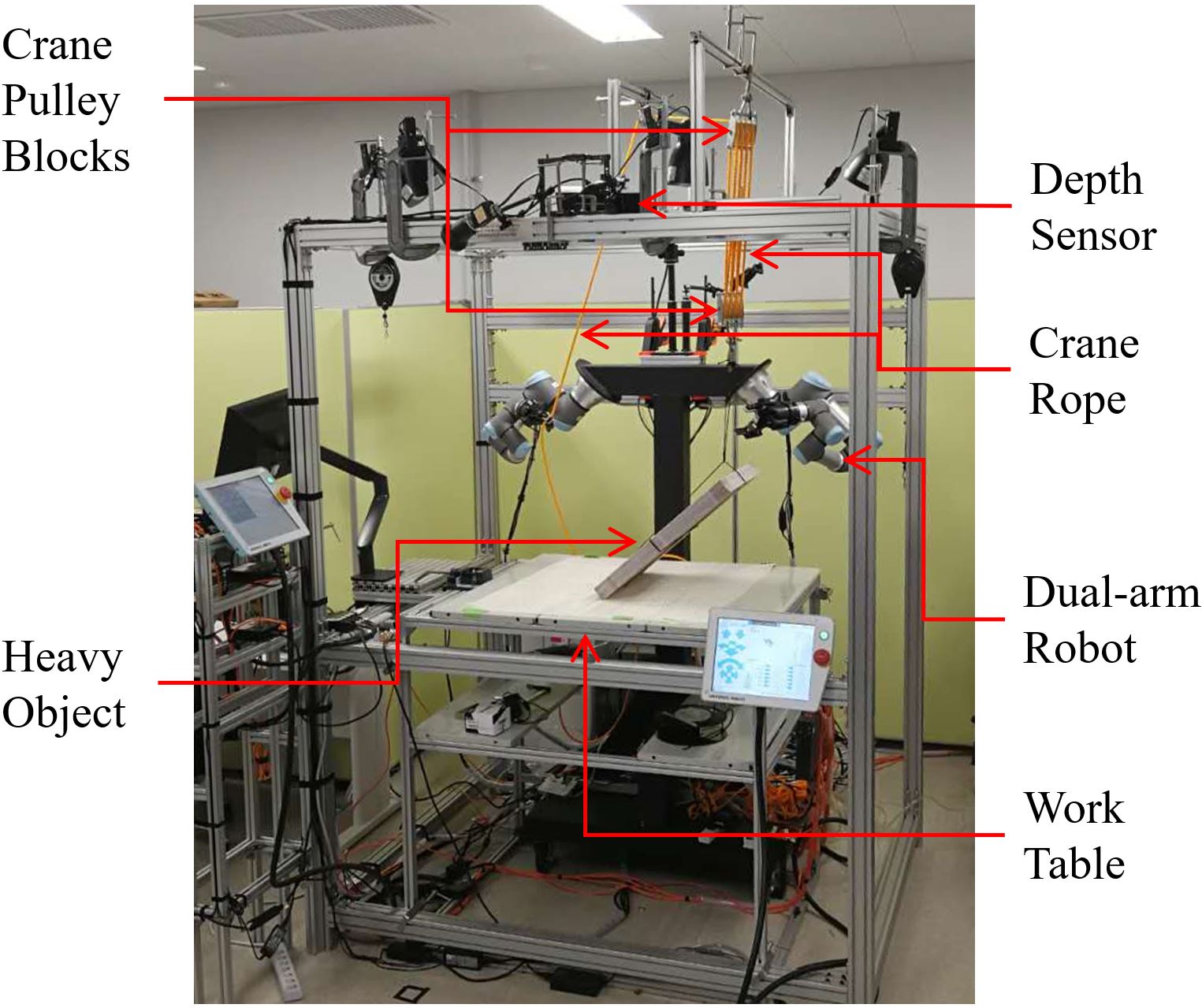}}
    \caption{An exemplary hardware setup. The goal is to optimize the robot action sequences and plan the robot motion that pulls up the heavy plate using the crane pulley blocks, and tumbles over the plate at a satisfying pose.}
    \label{hardware}
\end{figure}

This section presents an overview of our proposed method. We base our discussion on the hardware set up shown in Fig.\ref{hardware}. The proposed method is not limited to this hardware setup, but we use it as an example to let our readers have a solid conception of the working scenario.

The proposed method is divided into two phases. The first phase is a rope pulling phase. In the beginning this phase, the method detects a rope's position using point clouds obtained from a depth sensor. Many well-developed algorithms, for example, RANSAC \cite{raguram12} and ICP \cite{zhang94icp}, can be used to perform the detection. The detected rope, together with the pre-annotated grasp poses that grasp a section of the rope, is used to examine which robot arm to use and optimize the initial and goal rope-pulling poses. The two arms are examined sequentially, but they do not necessarily move one by one. The actuation of arms is determined by a quality function designed to maximize the pulling distance, minimize the pulling force, and enlarge grasping flexibility so that the plate object can be quickly and safely lifted up. An arm motion will be planned and executed to pull the crane rope if an optimal goal is found. After pulling the rope, the method performs a second visual detection to find the plate's pose. It determines if a plate is at a satisfying pose for tumbling. If the pose is not satisfying, the routine returns to the rope detection and pulling optimization to pull down the rope continuously. Or else, the method switches to a tumbling phase that optimize the trajectory of a robot arm that flips the plate using sliding-push. The tumbling is also monitored by vision to determine if the plate is well flipped or not. Once the plate's pose is considered to be reversed, the robot uses its two arms to return the crane rope and lower the plate down to the table.

The detailed algorithms related to the two phases will be presented in detail in Section IV and V, respectively. Their performance will be examined and analyzed in Section V.

\section{Pulling the Rope Using Optimized Poses}
\label{section4}
\begin{figure}[!htbp]
    \centering
    \includegraphics[width=.9\linewidth]{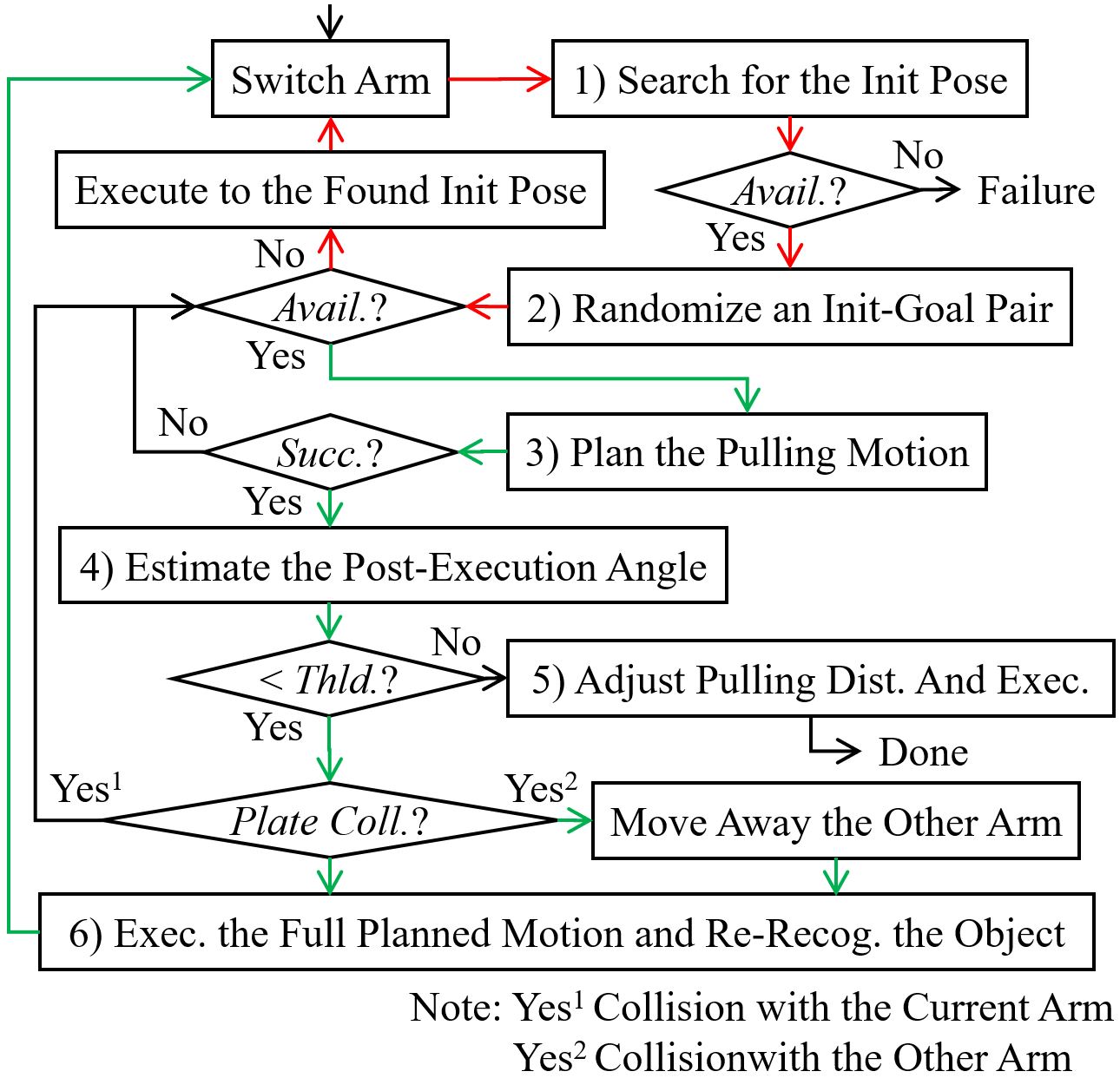}
    \caption{Workflow of the optimal rope-pulling planning algorithms. The planning is performed repeatedly in closed loops, as denoted by the red arrows and green arrows. Inside each loop, the robot uses a quality function to determine the best initial and goal pulling poses, and plans and executes the planned actions until a plate is lifted over a threshold angle.}
    \label{figpullflow}
\end{figure}

In this section, we present the optimal rope-pulling planning algorithms used in the rope pulling phase. The robot lifts up a plate by repeatedly carrying out the planned rope-pulling motion. As shown in Fig.\ref{figpullflow}, the planning is performed repeatedly in closed loops. Inside each loop, the robot uses a quality function to determine the best initial and goal pulling poses, and plans and executes the planned pulling actions until the plate is lifted up to a satisfying angle. If the planning or execution in a loop fails, the planner will invalidate the failed grasping point or pulling pose and try another candidate. The robot autonomously switches between single-arm or dual-arm actions following the determined goals.

The components of Fig.\ref{figpullflow} will be presented in detail in the remaining part of this section. Frame box 1) will be presented in subsection A. Frame box 2) will be presented in subsection B. Frame boxes 3)-5) will be presented in subsection C. The closed-loop arrows (the red arrows and the green arrows respectively form closed loops) and the switches of arms will be presented in subsections D. Frame box 6) will be presented in subsection E.

\subsection{Searching for the Initial Grasping Point and Pulling Poses}
\label{sub1}
\begin{figure*}[!htbp]
    \centering
    \includegraphics[width=.97\linewidth]{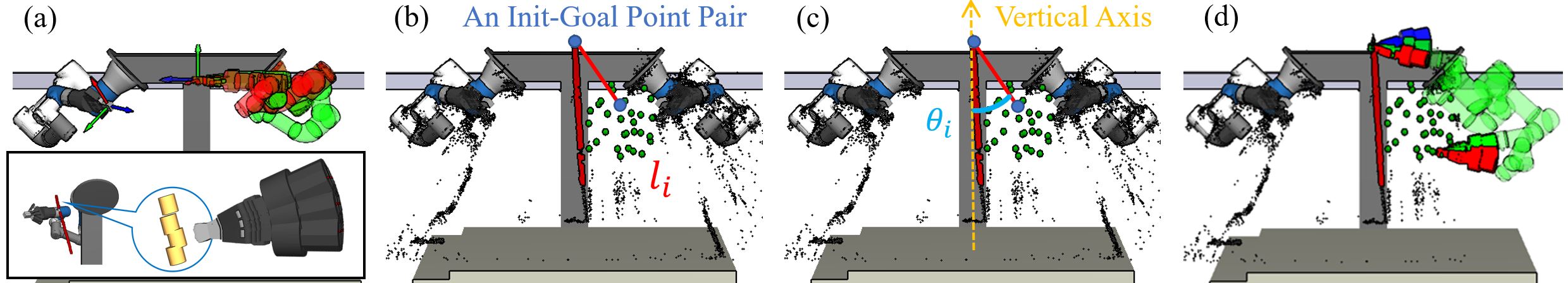}
    \caption{(a) Searching for the initial grasp point and pulling poses considering cylinder elements and pre-annotated grasp poses. The lower part of the figure shows the cylindrical modeling of the rope and the pre-annotated grasps for a cylinder element. (b, c, d) Sampling and determining an optimal init-goal pair considering the pulling distance computed using $l_i$, the pulling load computed using $\theta_i$, and the chance of successful motion planning computed by reasoning shared grasps. The random green dots are the sampled goals. The two clusters of hands in (d) show the grasping poses at the initial and goal points. The red and green ones are the shared grasping poses at both the initial and goal points. The blue ones in the upper cluster are not accessible at the goal. In the shared grasping poses, the red ones are IK-infeasible or collided. The green ones are the finally determined candidate grasping poses. Their related arm pulling poses are also rendered in green color.}
    \label{figinitgoal}
\end{figure*}
The initial grasping point and initial pulling poses for the rope-pulling motion are determined considering the rope's point cloud obtained inside each loop. We use a series of connected cylinders to model a detected rope point cloud, as shown in Fig.\ref{figinitgoal}(a). The cylinder elements have the same size. Grasp poses are pre-annotated for a cylinder element. To determine initial pulling poses, we scan the cylinder elements of a detected rope from its top-most position and select the first one that is not invalidated by previous loops as the initial grasping point. Then, the pose of an arm is computed by solving Inverse Kinematics (IK) considering the pre-annotated grasp poses at the selected point. After that, the collision-free ones of the solved arm poses are saved as the initial pulling poses. The green rendering in the upper part of Fig.\ref{figinitgoal}(a) exemplifies a determined initial pulling pose. The red rendering shows an obsoleted grasping pose as the arm elbow at this configuration collides with the robot body.

\subsection{Determine the Optimal Init-Goal Pair}
\label{sub2}
After the initial grasping point and pulling poses are decided, the planner continues to decide goal points and the pulling poses at them to form init-goal pairs. The goal points are selected from a set of probabilistically sampled points in a robot arm's workspace, as shown in Fig.\ref{figinitgoal}(b-d). Since the two arms have different workspaces, their goal points are sampled differently. The most effective point in the sampled set is selected to form the initial-goal pair, where effectiveness is evaluated considering three aspects: The motion distance between the init-goal pair; The load that a pulling arm bears when moving between the init-goal pair; The number of shared grasp poses between the init-goal pair. The three aspects are respectively quantified as quality parameters $f_{length}$, $f_{load}$, and $f_{grasps}$. The values of these quality parameters are normalized to (0, 1). Their details are as follows.

\subsubsection{$f_{length}$}
The $f_{length}$ quality assesses the motion distance between the init-goal pair. Its goal is to enlarge the length of each pulling motion. $f_{length}$ is computed using equation \eqref{eq:f_length}.
\begin{equation}
    f_{length} = (l_{i}-l_{min}) / (l_{max} - l_{min}).
    \label{eq:f_length}
\end{equation}
Here, $l_i$ is the length between the upper pin point of the pulley blocks and a sampled goal point. It is graphically explained in Fig.\ref{figinitgoal}(b). $l_{max}$ and $l_{min}$ are respectively the shortest and largest of all $l_{i}$. The denominator $(l_{max} - l_{min})$ normalizes the values. $f_{length}$ reaches to 1 when the furthest goal sample is selected.

\subsubsection{$f_{load}$} The $f_{load}$ quality assesses the load that a pulling arm bears when moving between the init-goal pair. Its goal is to reduce the load of a pulling arm. The $f_{load}$ value is computed using the tilting angle of the stretched rope. A Free Body Diagram (FBD) analysis shows that a larger tilting angle requires a robot arm to afford larger forces. Thus we use the cosine value of the angle to approximate $f_{load}$:
\begin{equation}
    f_{load} = \cos{\theta_{i}}
    \label{eq:f_FT}
\end{equation}
Fig.\ref{figinitgoal}(d) graphically explains $\theta_{i}$. $f_{load}$ reaches 1 when the rope is pulled vertically. The robot arm bears the smallest resistant force at this extreme.

\subsubsection{$f_{grasps}$} The $f_{grasps}$ quality aims to improve the chances of successful motion planning. It is computed using the number shared grasp poses at the initial grasping point and goal point as follows:
\begin{equation}
    f_{grasps} = n_{goal} / n_{init}.
    \label{eq:f_manip}
\end{equation}
A shared grasp pose is defined as the grasp poses identical in a cylinder element's local coordinate frame at both the initial grasping point and the goal point. Shared grasps are computed by reasoning the intersection of all available (IK-feasible and collision-free) grasp poses at the init-goal pair. $n_{init}$ in the equation indicates the number of available grasp poses at the initial grasping point. $n_{goal}$ indicates the number of grasp poses at the goal point that shares the same local transformation as those in $n_{init}$. A large $f_{grasps}$ means more candidate initial and goal pulling poses for following configuration-space motion planning, which implies a higher chance of getting a feasible motion trajectory.

The most effective goal point is determined considering the three quality parameters using the following quality function.
\begin{align}
    Q & = \boldsymbol{\omega}^T \boldsymbol{f} \label{eq:J} 
\end{align}
Here, $\boldsymbol{\omega}$=[$\omega_{length}$, $\omega_{load}$, $\omega_{grasps}$] is the weighting coefficient. $\boldsymbol{f}$=[$f_{length}$, $f_{load}$, $f_{grasps}$] indicates the three quality parameters. The quality function helps to find a goal point that leads to large pulling distances, smaller pulling forces, and high chance of finding successful pulling motion. The weighting coefficient can be tuned following application requirements. Their influences on final performance will be analyzed in the experimental section. 

For all sampled goal points, we compute their $Q$ and choose the one with the largest $Q$ value as the optimal goal. The initial grasping point and the optimal goal point together form an init-goal point pair. The goal arm pulling poses are determined by considering the collision-free Inverse Kinematics (IK) solutions of the shared grasp poses at the init-goal pair. The two clusters of hands in Fig.\ref{figinitgoal}(d) show the grasping poses at the initial and goal points. The blue ones in the upper cluster indicate the unshared grasping poses. They are inaccessible at the goal. The ones with the other colors indicate the shared grasping poses. In the shared grasping poses, the red ones are IK-infeasible or collided. The green ones are the finally determined candidates. Their related arm pulling poses are rendered for better observation.

\subsection{Planning the Rope-Pulling Motion}
\label{sub3} \par
\begin{figure}[!htbp]
  \begin{minipage}[c]{0.64\linewidth}
    \includegraphics[width=\textwidth]{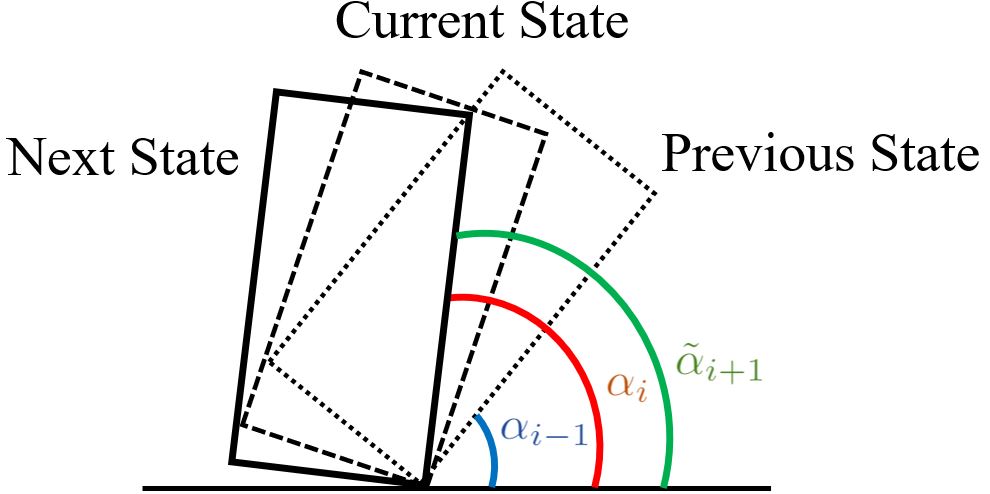}
  \end{minipage}~
  \begin{minipage}[c]{0.31\linewidth}
    \caption{Predicting a plate's next tilting angle $\Tilde{\alpha}_{i+1}$ using the current titling angle $\alpha_{i}$ and previous tilting angle $\alpha_{i-1}$.} 
    \label{fig:predictlist}
  \end{minipage}
\end{figure}

 
After the initial and goal pulling poses are determined, the robot generates a motion to pull a rope from the initial pose to the goal pose. The motion includes two sections. In the first section, the pulling arm moves from its current pose to the initial pulling pose. The section is planned using Bi-directional Rapidly-exploring Random Trees (Bi-RRT) \cite{berenson2009manipulation}. In the second section, the pulling arm pulls the rope by moving from the initial pulling pose to the goal pulling pose. The section is planned as a linear trajectory using Quadratic Programming (QP) \cite{bobrow2001optimal}\cite{harada2006natural}. If the planning succeeds, the robot will execute the successfully planned motion to pull up the plate. Or else, the planner will continue to try another init-goal pair. The planning and execution routine is performed repeatedly until the plate's tilting angle exceeds a threshold. We use a threshold because we need to ensure that the plate is not lifted too much and separated from the table. The plate continuously contacts with the table until the end of the rope-pulling phase, so that a robot arm could start tumbling manipulation. Before the execution in each loop, the plate's post-execution tilting angle is predicted to prevent it from exceeding the threshold. The post-execution angle prediction is performed using the following equation.
\begin{align}
    \Tilde{\alpha}_{i+1} & = \alpha_{i} + (\alpha_{i} - \alpha_{i-1}) \times \frac{d_{i}}{d_{i-1}},~(i\geq 1)
    \label{eq:predictequation}
\end{align}
Here, $\alpha_{i-1}$ and $d_i$ are respectively the previous plate's tilting angle and the length of the previous rope-pulling motion. $\alpha_{i}$ and $d_i$ are the current plate's tilting angle and the length of the current rope-pulling motion. The plate's tilting angle after executing the current rope-pulling motion is estimated as $\Tilde{\alpha}_{i+1}$. The various symbols are graphically explained in Fig.\ref{fig:predictlist}. The prediction is based on the proportional relationship between the length of the pulled rope and the change of the angle in a previous execution. It is decoupled from a specific plate and provides an upper-bound estimation for the post-execution angle. 

If the predicted angle exceeds the threshold, the length of the next pulling motion will be adjusted to avoid over-lifting. The following equation shows the adjustment. The plate's tilting angle after finishing the rope-pulling phase approximates to the threshold angle $\alpha_{thld}$. 
\begin{align}
    \hat{d}_{i} & = d_{i-1} \times \frac{\alpha_{thld} - \alpha_{i}}{\alpha_{i} - \alpha_{i-1}}
    \label{eq:changepullinglength}
\end{align}

\subsection{Determining the Action Arms and Sequences}
As mentioned in \textit{C}, the planning and execution loop is performed repeatedly until the plate's tilting angle exceeds a threshold. The two arms share the repetition sequentially -- The planner optimizes init-goal pairs and plans and executes a pulling-motion for the two arms one by one. It switches to a second arm either after pulling the cable or without any pulling action. Particularly, if no init-goal pair is found or no successful motion is planned, the planner switches to the next arm without pulling, exhibiting autonomy in determining the action arms and sequences.

The arrows in Fig.\ref{figpullflow} show the mentioned switches. The green arrows indicate the flow with pulling executions. The current arm along the flow is actuated to pull the cable. It will switch to the other arm after execution. The red arrows indicate the flow without pulling. During the planning, a failure may be caused by different reasons like no available initial pose, no available init-goal pair, failed to find a motion for an init-goal pair, etc. If the reason is a collision between the lifted plate and the other arm, the planner will try moving the other arm away and continue the currently planned results (as shown by ``Yes$^2$'' in Fig.\ref{figpullflow}). Or else, the planner will jump over the current init-goal pair and perform replanning by iterating to other candidates (as shown by the ``No'' condition after ``$Succ.$'' and the ``Yes$^1$'' condition after ``$Plate Coll.$'' in the figure). If none of the sampled goals form a feasible init-goal pair, the current arm will not perform the pulling action. Instead, it simply re-grips by opening the gripper, moving to the init pulling pose, and closing the gripper. The planner switches to plan for the other arm after the re-gripping. A final failure is reported when all cylinder elements on a detected rope are traversed, and none of them lead to a feasible initial pulling pose.

Following the green and red workflows, the action arms and sequences are determined autonomously. In an extreme case, a robot may use a single arm to pull while never find feasible goals for the other arm. The robot may also use the two arms one by one in another extreme case where both arms find feasible init-goal pairs. More moderate cases are that the robot sometimes uses a left arm and sometimes uses a right arm, depending on the amount of feasible init-goal pairs found during the repeated planning. 

\begin{figure}[!htbp]
    \centering
    \includegraphics[width=.85\linewidth]{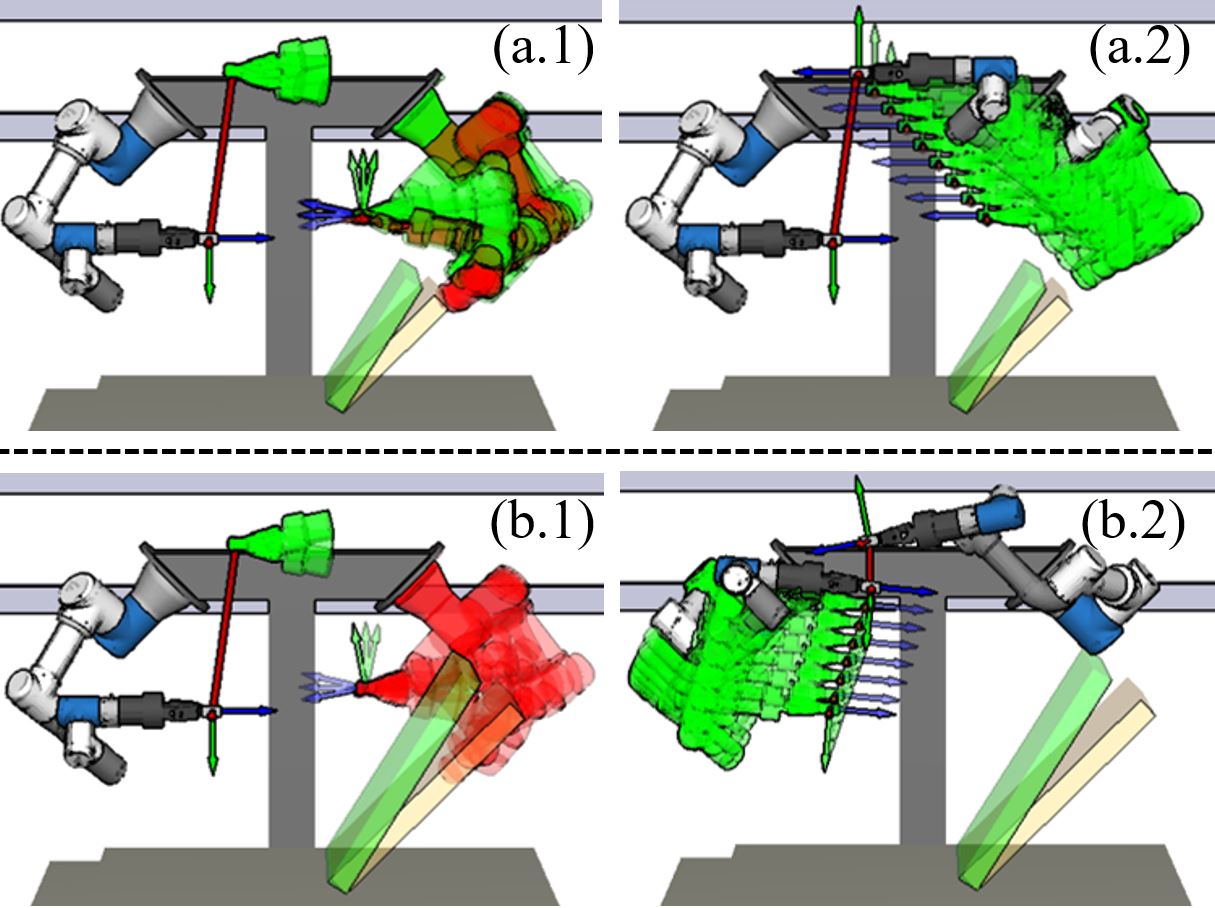}
    \caption{Two examples of switching the action arms. The goal is to lift the board from the yellow pose to the green pose. (a) The robot successfully finds an init-goal pair for its left arm and plans a pulling motion. (b) The robot fails to find a goal for its left arm because of the bulky plate's obstruction. It switches to the right arm without execution.}
    \label{fig:pullingpatern}
\end{figure}

Fig.\ref{fig:pullingpatern} shows two examples. In Fig.\ref{fig:pullingpatern}(a), the robot is pulling up a small plate. The robot finished a right-arm execution in (a.1) and is checking the collisions of the shared left-arm grasping poses. The top-most cylinder element on the detected rope is chosen as the init point. The rendered left arms are the shared IK-feasible poses. There are both collided and collision-free ones. The collided poses are shown in red, and the collision-free poses are shown in green. In (a.2), the robot plans a linear motion to move the left arm from an initial pulling pose to a goal pulling pose. Fig.\ref{fig:pullingpatern}(b) shows the case of larger plate. All shared IK-feasible grasp poses of the left arm collide with the plate in (b.1). The planner reaches to a final failure and switches to the right arm without triggering execution. In (b.2), the right arm finds a new init-goal pair and plans a linear motion to move the right arm from an initial pulling pose to a goal pulling pose.

\subsection{Include Re-Recognition in the Loop}
\begin{figure}[!htbp]
    \centering
    \includegraphics[width=.95\linewidth]{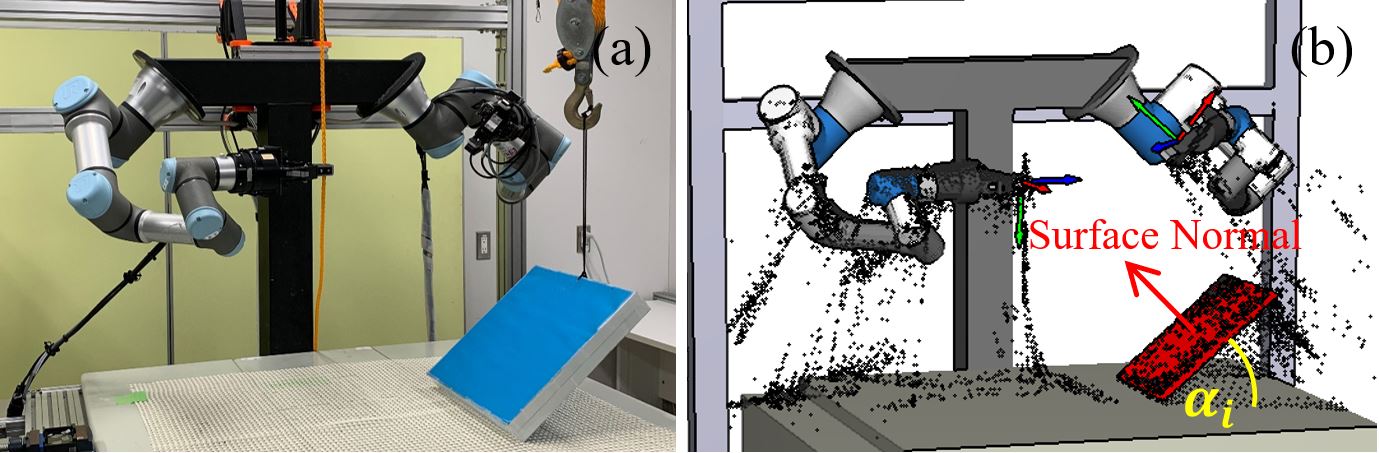}
    \caption{A plate's real tilting angle is re-recognized using a depth camera after each execution. The algorithm extracts the largest planar surface and computes the real tilting angle considering the normal of the extracted surface.}
    \label{figest}
\end{figure}

The plate pose is re-recognized using a depth camera after each execution to acquire its real tilting angle. Like recognizing a rope, RANSAC and ICP are used to extract the plate's largest planar surface. The real tilting angle of the plate is computed considering the normal of the extracted planar surface. It is then used to update the $\alpha_{i}$ in equation (\ref{eq:predictequation}) as well as to predict the $\Tilde{\alpha_{i}}$ of the next rope-pulling action. Fig.\ref{figest} shows an example. The point cloud of the scene in (a) is acquired and shown in (b). The estimated planar surface normal and real tilting angle are illustrated together with the point cloud. The optimization, planning, execution, and recognition workflows mentioned in this section's subsections are repeated for the two arms sequentially until the plate is lifted up to a given threshold angle.

\section{Tumbling the Plate Using Sliding-Push}
\label{sec:pushing}
After lifting the plate to the threshold angle, one robot arm will tumble it over by pushing. Here, we assume that a plate always has edge contacts with the table. It will be tumbled by rotating around the edge contacts. The rotation trajectory of a plate is modeled as a time-variant function, and the robot arm follows the trajectory by using sliding-push. We define a sliding-push as a pushing policy that allows sliding on the contact surface of a plate. Like the rope-pulling actions, the sliding-push motion is performed continuously until the plate is rotated beyond a threshold pose where the force and moment reach equilibrium when the robot moves away from the pushing point. 

Besides the edge contacts, we also assume the following conditions and constraints in planning the sliding-push: (1) All forces appear inside a vertical plane perpendicular to the contact edges; (2) The plate and the environment are rigid; (3) The push between two nearby time instants is quasi-static; (4) The friction at the contact point follows the Coulomb friction model; (5) Plate-slipping on the table surface is not considered, but finger-sliding on the contact surface is allowed. Especially the fifth assumption is a strong constraint and leads to fewer solutions, but we keep it there to narrow down the search space. A plate has three contacts during pushing -- Rope connection, robot fingertip contact, and table contact. The force at the connecting point with the rope is difficult to control. The contact between the fingertip and the plate is assumed to be a point contact. During pushing, the plate may both slide on the table surface and rotate around the contact edge, making it extremely difficult to perform optimization. To avoid these problems, we develop the fifth assumption to constrain a robot arm rotating a plate without allowing it to slide on the table surface. Pushing actions will be carefully optimized considering the constraint about table-slipping. Meanwhile, we allow a pushing finger to slide on the plate's contact surface to maintain high success rate. 

\begin{figure}[!htbp]
    \centering
    \includegraphics[width=.87\linewidth]{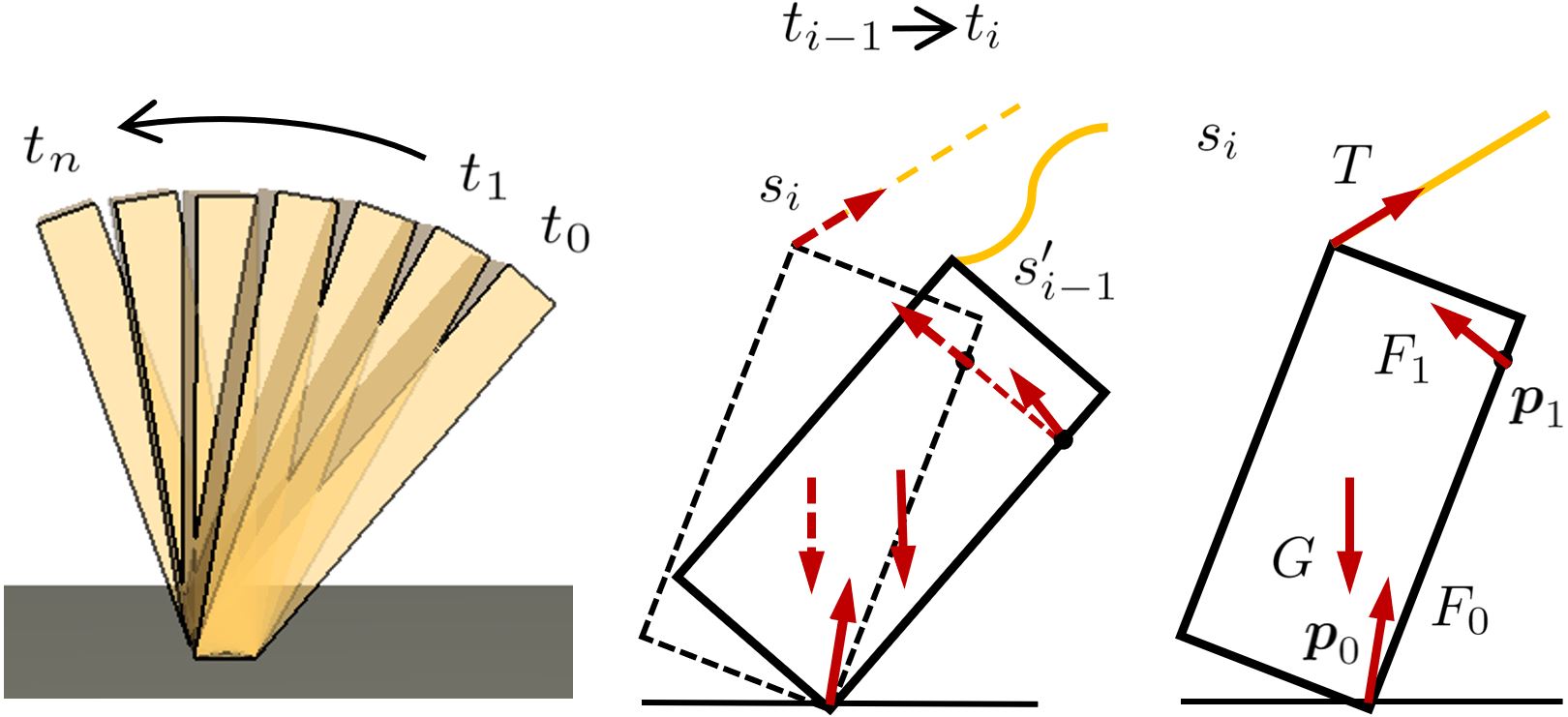}
    \caption{The rotational trajectory of a plate is modeled as a time-variant function across $t_0$, $t_1$, ..., $t_n$. A pushing at a time instant $t_i$ is divided into two states $s_i$ and $s_i^\prime$. The goal of optimization is to simultaneously minimize $F_0$, $F_1$, and $T$, while considering a changing $\boldsymbol{r}_1=\protect\overrightarrow{\boldsymbol{p}_0\boldsymbol{p}_1}$ (sliding-push).}
    \label{figtumblingsyms}
\end{figure}

Based on these assumed constraints, a plate's motion trajectory is determined once the contact edge is known. The tumbling action is implemented by optimizing a robot arm's pushing trajectory while considering the plate's motion trajectory. As the pushing arm tumbles a plate, the remaining robot arm holds or loosens the crane rope alternatively to avoid jerks. Take the plate pose at a time instant $t_i$ shown in Fig.\ref{figtumblingsyms} for example. We divide the pushing at this time instant into two states $s_{i}$ and $s_{i}^\prime$. The plate is pushed from a previous pose at $t_{i-1}$ to the current pose in the first state. The crane rope is straightened, and a tension force $T$ appears along the rope. In the second state, the rope is loosened by the other robot arm, and the rope tension $T$ disappears. The tumbling is performed across a sequence of these states at different time instants in the following way.
\begin{equation}
    \underbrace{s_0\xrightarrow{lsn} s_0^\prime}_{t_0} \xrightarrow{hld} \underbrace{s_1 \xrightarrow{lsn} s_1^\prime}_{t_1} \xrightarrow{hld} \ldots \underbrace{s_{n-1} \xrightarrow{lsn} s_{n-1}^\prime}_{t_{n-1}} \xrightarrow{hld} \underbrace{s_n}_{t_n}
    \nonumber
\end{equation}
Between each $s_{i-1}^\prime$ and $s_{i}$, the other arm holds ($hld$ for abbreviation) the crane rope. Between each $s_{i}$ and $s_{i}^\prime$, the crane rope is loosened ($lsn$ for abbreviation). We analyze the forces at each time instant and determine the pushing points at each of them by minimizing the balancing forces and reducing the rope tension. The formal expressions for the minimization is as follows. 

\begin{subequations}\allowdisplaybreaks
    \begin{align}
    &\underset{\boldsymbol{r}_1}{\text{min}} & \nonumber{k_1 (F_0^{T}F_0)}+{k_2(F_1^{T}F_1)}+{k_3(T^TT)} && \\ \label{opgoal}\\
    &\text{s.t.} & T + G + \sum_{j=0} F_j = 0 \label{opsif} \\
    && \boldsymbol{r}_{t} \times T + \boldsymbol{r}_{g} \times G + \sum_{j=0} \boldsymbol{r}_{j} \times F_{j} = 0 \label{opsit}\\
    && F_1^TF_1 \in (0, 30) \label{opsif1c}\\
    && \frac{f_{0x}}{f_{0y}} \in (0, \mu_0)\label{opsifriction0c}\\
    && \mathtt{acos}(\frac{F_1\cdot \boldsymbol{r}_1}{||F_1||||\boldsymbol{r}_1||}) \in (\mathtt{atan}\frac{1}{\mu_1}, \pi-\mathtt{atan}\frac{1}{\mu_1})\label{opsifriction1c}\\
    && \boldsymbol{r}_1(t_{i})^T\boldsymbol{r}_1(t_{i}) \in (0, l)\label{opsrng}\\
    && |\boldsymbol{r}_1(t_{i})-\boldsymbol{r}_1(t_{i-1})| \leq |\boldsymbol{v}_{max}|(t_i-t_{i-1})\label{opsspeed} \\
    && \mathtt{acos}(\frac{(\boldsymbol{r}(t_{i+1})-\boldsymbol{r}(t_{i}))\cdot (\boldsymbol{r}(t_{i})-\boldsymbol{r}(t_{i-1}))}{||\boldsymbol{r}(t_{i+1})-\boldsymbol{r}(t_{i})||||\boldsymbol{r}(t_{i})-\boldsymbol{r}(t_{i-1})||}) \leq \gamma \label{oppushdir}
    \end{align}
\end{subequations}

The meanings of the various variables are listed below. They are also graphically illustrated in Fig.\ref{figtumblingsyms} for readers' convenience.
\begin{itemize}
\item[$F_0$] Force at the table contact point $\boldsymbol{p}_0$ in an $s_i$ state. An edge contact is simplified into a point contact. $F_0$ = [$f_{0x}$, $f_{0y}$].
\item[$F_1$] Force at the robot finger tip contact point $\boldsymbol{p}_1$ (pushing point) in an $s_i$ state. $F_1$ = [$f_{1x}$, $f_{1y}$].
% \item[$F_0^\prime$] Force at the table contact point $\boldsymbol{p}_0$ in an $s_i^\prime$ state. Like $F_0$, only a point contact is considered. $F_0$ = [$f_{0x}$, $f_{0y}$].
% \item[$F_1^\prime$] Force at the robot finger tip contact point $\boldsymbol{p}_1$ (pushing point) in an $s_i^\prime$ state. $F_1$ = [$f_{1x}$, $f_{1y}$].
\item[$G$] Gravity.
\item[$T$] Tension caused by a stretched crane rope. Only exist at an $s_i$ state.
\item[$\boldsymbol{r}_t$] The vector pointing from the rotation center to the rope connecting point.
\item[$\boldsymbol{r}_g$] The vector pointing from the rotation center to the plate's center of mass.
\item[$\boldsymbol{r}_{j}$] The vector pointing from the rotation center to $\boldsymbol{p}_i$.
\item[$\mu_0$] The Coulomb friction coefficient at $\boldsymbol{p}_0$.
\item[$\mu_1$] The Coulomb friction coefficient at $\boldsymbol{p}_1$.
\item[$\boldsymbol{v}_{max}$] The maximum speed of a pushing arm.
\item[$(t_i)$] If a symbol does not have this postscript, it always denotes a value at time instant $t_i$. Or else, the symbol denotes a value from the time instant shown in the parenthesis.
% \item[$l$] The length of the contact surface for pushing.
% \item[$s_{i}$] The direction vector of pushing from $s_{i}$ to $s_{i+1}$
\item[$\gamma$] The maximum allowable angle between current and previous pushing directions.
\end{itemize}

\begin{figure}[!htbp]
    \centering
    \includegraphics[width = .85\linewidth]{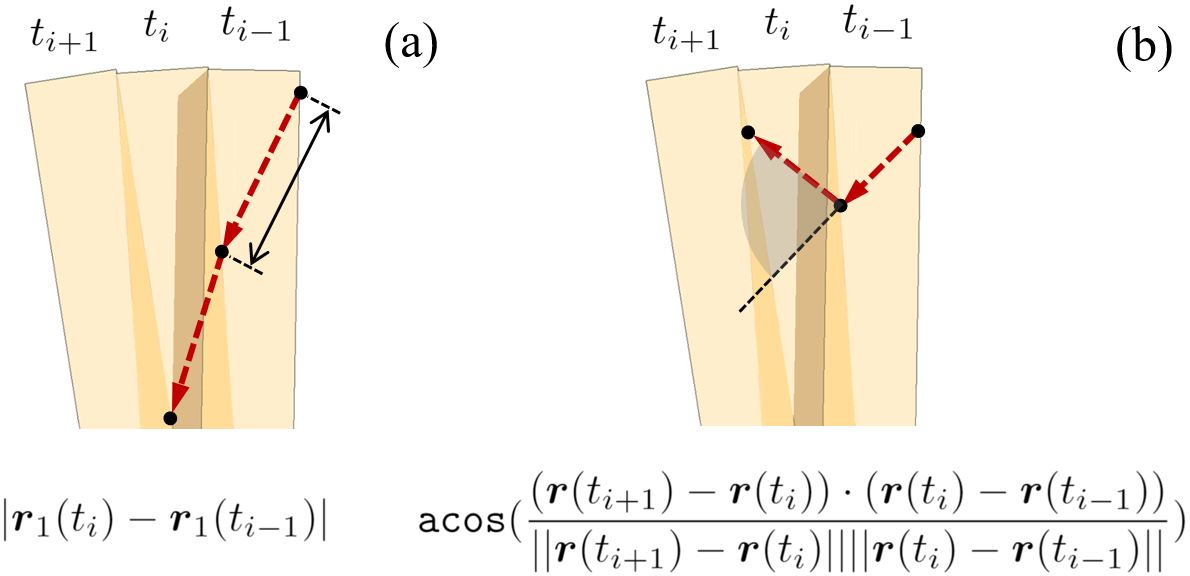}
    \caption{(a) Equation \eqref{opsspeed} constrains the maximum distance between adjacent pushing positions. It makes the blue arrows in the figure short. (b) Equation \eqref{oppushdir} constrains the deviation of a subsequent pushing direction with its precedence. It helps to avoid oscillation and make the trajectory smooth.}
    \label{fig:pushcons}
\end{figure}

The optimization goal show in equation \eqref{opgoal} is to minimize the balancing force needed in the entire system. The idea behind this optimization is that by minimizing the force applied to the entire system, the robot can push the plate with a minimum force and reduce the external forces from the rope and the desk. The smaller forces will also decrease the risk of rotating in an unexpected direction (caused by non-vertical rope tension) or slipping on the table. The constraints \eqref{opsif}, \eqref{opsit} balance the forces and torques at an $s_i$ state. The constraints \eqref{opsif1c}, \eqref{opsifriction0c}, \eqref{opsifriction1c} add bounds to the force exerted by a robot and the friction cone at the contact points at the $s_i$ state. The constraint \eqref{opsrng} limits $\boldsymbol{r}_1$ to be on a contact surface. The constraint \eqref{opsspeed} ensures the contact points at two consecutive time instants are reachable. The last constraint \eqref{oppushdir} smooths the change in the pushing direction. It prevents the pushing direction at a time instant from getting largely diverted from its precedence. Both constraints \eqref{opsrng} and \eqref{oppushdir} are important to finding a practical pushing trajectory. Their roles and difference are illustrated in Fig.\ref{fig:pushcons}. Essentially, the constraint \eqref{opsrng} limits the pushing distance between adjacent pushing points. The constraint \eqref{oppushdir} limits the changes in pushing directions.

The minimization is performed across all $s_0$, $s_0^\prime$, $s_1$, $s_1^\prime$, ... states to determine a sequence of optimal $\boldsymbol{r}_1$, say an optimal sliding-pushing trajectory. Note that during the optimization the robot kinematic constraint is not considered. Instead of including it integrally as a constraint, we examine the kinematic constraint lazily after a sequence is found. If the robot IK is not solvable, we switch to the next best $\boldsymbol{r}_1$ until success or a failure is reported. With the IK constraints, the optimization will find both a sequence of pushing points and a sequence of robot joint trajectories that produce the pushing motion along with the pushing points. The joint trajectories will be executed by the target arm. The other arm plays the role of holding and loosening the crane rope following the changes of states. As mentioned at the beginning of this section, the resulted motion is a sliding-push. The contact points at different $t_i$ are not necessarily the same in a plate's local coordinate system. The pushing finger may slide on the contact surface over consecutive $t_i$ during the tumbling.

\begin{figure}
    \centering
    \includegraphics[width=.93\linewidth]{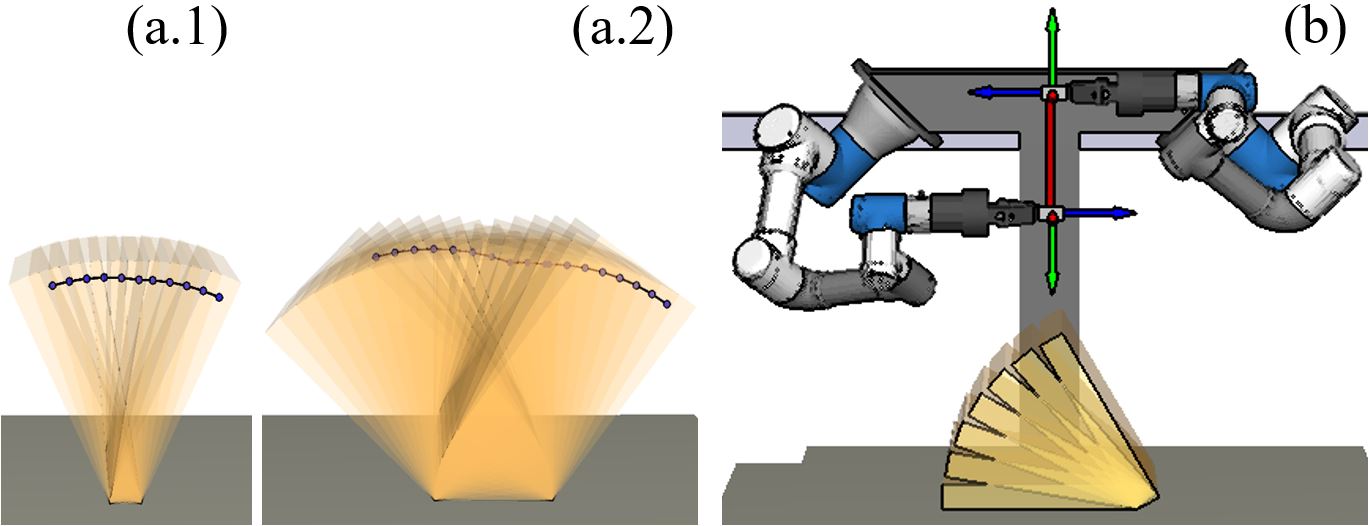}
    \caption{(a.1-2) The optimized trajectories experience slight changes as the plate rotates onto a second contact edge. The change gets significant as the plate becomes thick. (b) Returning the rope using both arms to completely lower down a plate onto the table.}
    \label{figdoubleedge}
\end{figure}

In addition, the contact edge between a plate and a table surface may change during the tumbling as a plate always has a thickness. Our tumbling planner takes into account the changes of the contact edge and optimizes the pushing trajectory considering the changing rotating axes. Fig.\ref{figdoubleedge}(a.1-2) show the found pushing trajectories for two plates with different thickness. The trajectories experience slight changes as the plate rotates onto a second contact edge.

Finally, after the pushing robot arm finishes the whole planned trajectory, it works together with the other arm to return the rope and to further lower down the plate on the table. Fig.\ref{figdoubleedge}(b) illustrates the returning actions. They are also planned online with visual rope detection. The plate's real tilting angle is monitored during returning to make sure it is completely lowered onto the table. The difference is there is no optimization, and the robot moves along a straight line pointing to the upper pulley block. The full task is reported to have been completed after the plate reaches the table surface.

%%%%%%%%%%%%%%%%%%%%%%%%%%%%%%%%%%%%%%%%%%%%%%%%%%%%%%%%%%%%%%%%%%%%%%%%%%%%%

\section{Experiments and Analysis}
\label{section5}
The proposed method is implemented and examined using the robot system shown in Fig.\ref{hardware}. The system includes two UR3e robots with two Robotiq F85 two-finger grippers at each robot end flange. A Kinect V2 (Microsoft) sensor is used to acquire 3D point clouds. The computer used for planning and optimization is a PC with Intel Core i9-9900K CPU, 32.0GB memory, and GeForce GTX 1080Ti GPU. The programming language for implementing the algorithms is Python.

\subsection{Influence of the Goal Quality Function for Pulling}
First, we analyze the influence of the quality function presented in equation (\ref{eq:J}), and examine the changes of the selected pulling goals under different weights. We perform the analysis by setting one of the three weights to a higher value while keeping the other two to zero and observe the planned results. Specifically, we compare $\omega$ = [$\omega_{length}$, $\omega_{load}$, $\omega_{grasps}$] = $[1, 0, 0]^T$, $[0, 1, 0]^T$, and $[0, 0, 1]^T$ and show the distribution of the selected goals for the left and right arms respectively.

\subsubsection{$\omega$ = $[1,0,0]^T$}
\begin{figure}[!htbp]
    \centering
    \includegraphics[width=\linewidth]{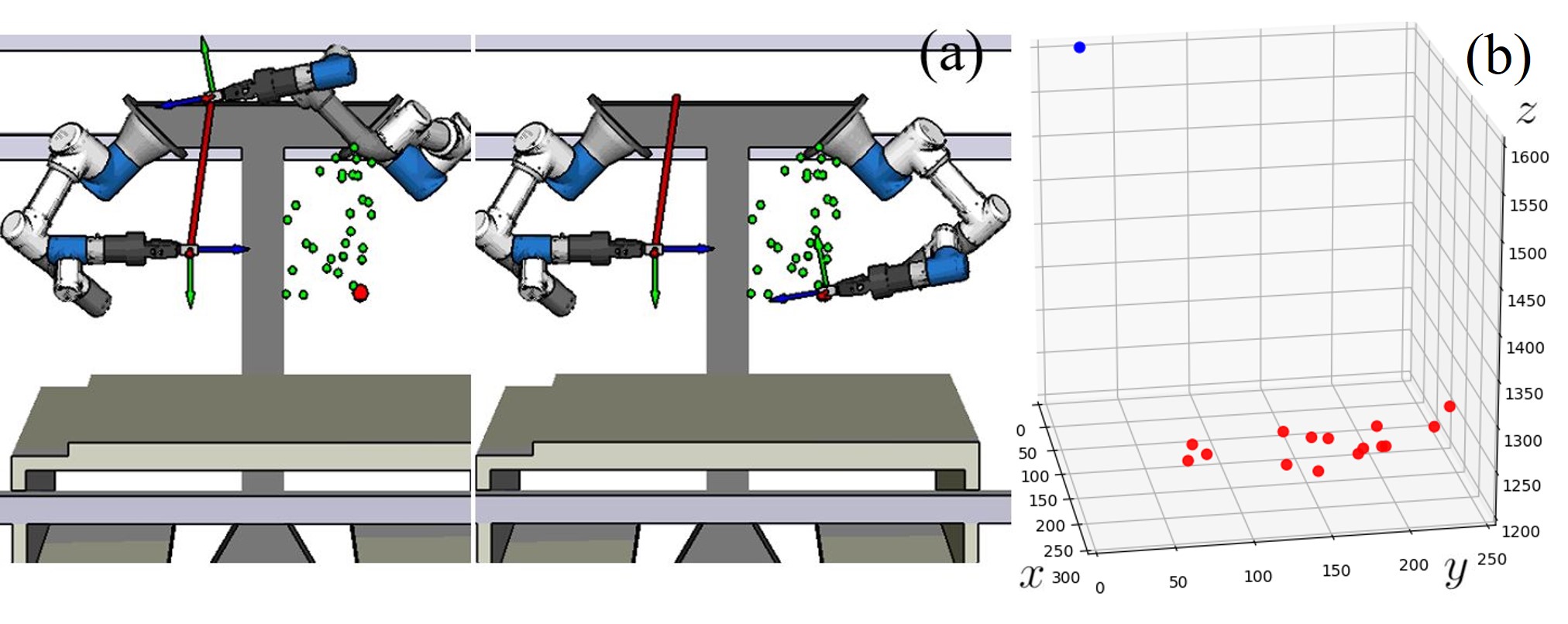}
    \caption{Results when $\omega$ = $[1,0,0]^T$: (a) An example of a selected init-goal pose pair. (b) A distribution of the selected goals under 15 times of simulation.}
    \label{flen}
\end{figure}

In this case, only $f_{length}$ influences the rope-pulling motion. It drives the robot to select goal points that lead to a large pulling distance. Fig.\ref{flen}(a) and (b) show an example of a selected init-goal pose pair and a distribution of the selected goal points under the weight setting, respectively. The robot selects faraway points for both the left and right arms to pull the rope with long distances.

\subsubsection{$\omega$ = $[0,1,0]^T$}
\begin{figure}[!htbp]
    \centering
    \includegraphics[width=\linewidth]{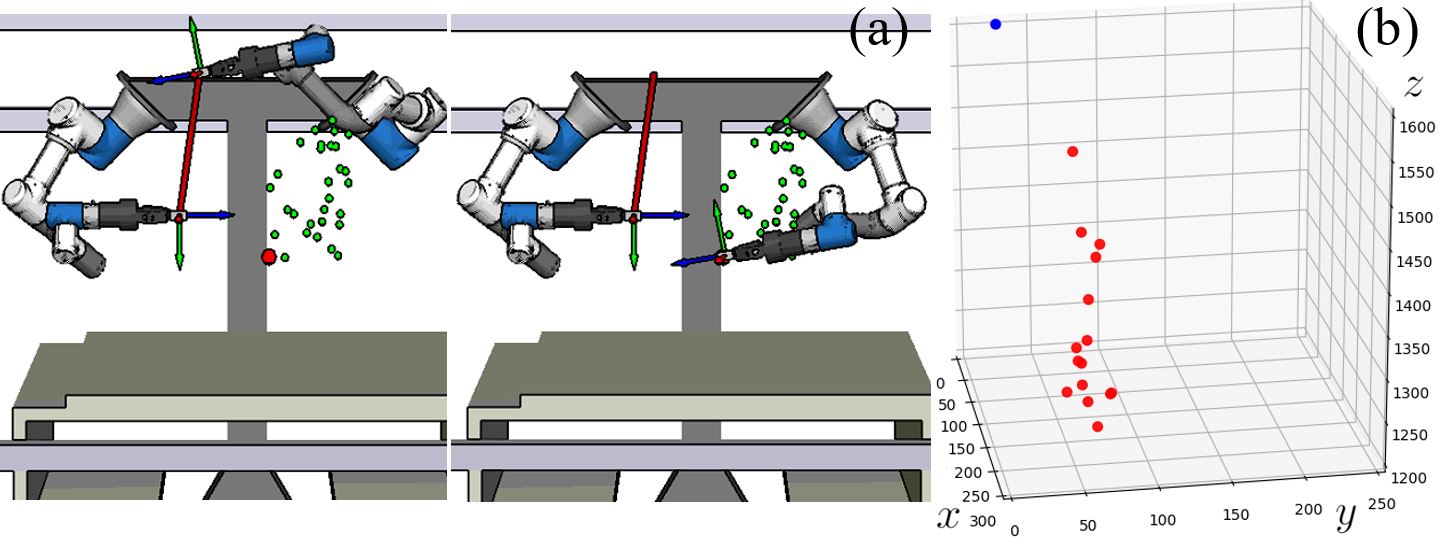}
    \caption{Results when $\omega$ = $[0,1,0]^T$: (a) An example of a selected init-goal pose pair. (b) Distribution of the selected goals after 15 times of simulation.}
    \label{fload}
\end{figure}

In this case, only $f_{load}$ affects the pulling motion. Fig.\ref{fload} exemplifies an init-goal pair and a statistical distribution of the selected goal points. The robot tends to select goals that make the stretched rope to have small tilting angles to reduce the forces that an arm needs to bear.

\subsubsection{$\omega$ = $[0,0,1]^T$}
\begin{figure}[!htbp]
    \centering
    \includegraphics[width=\linewidth]{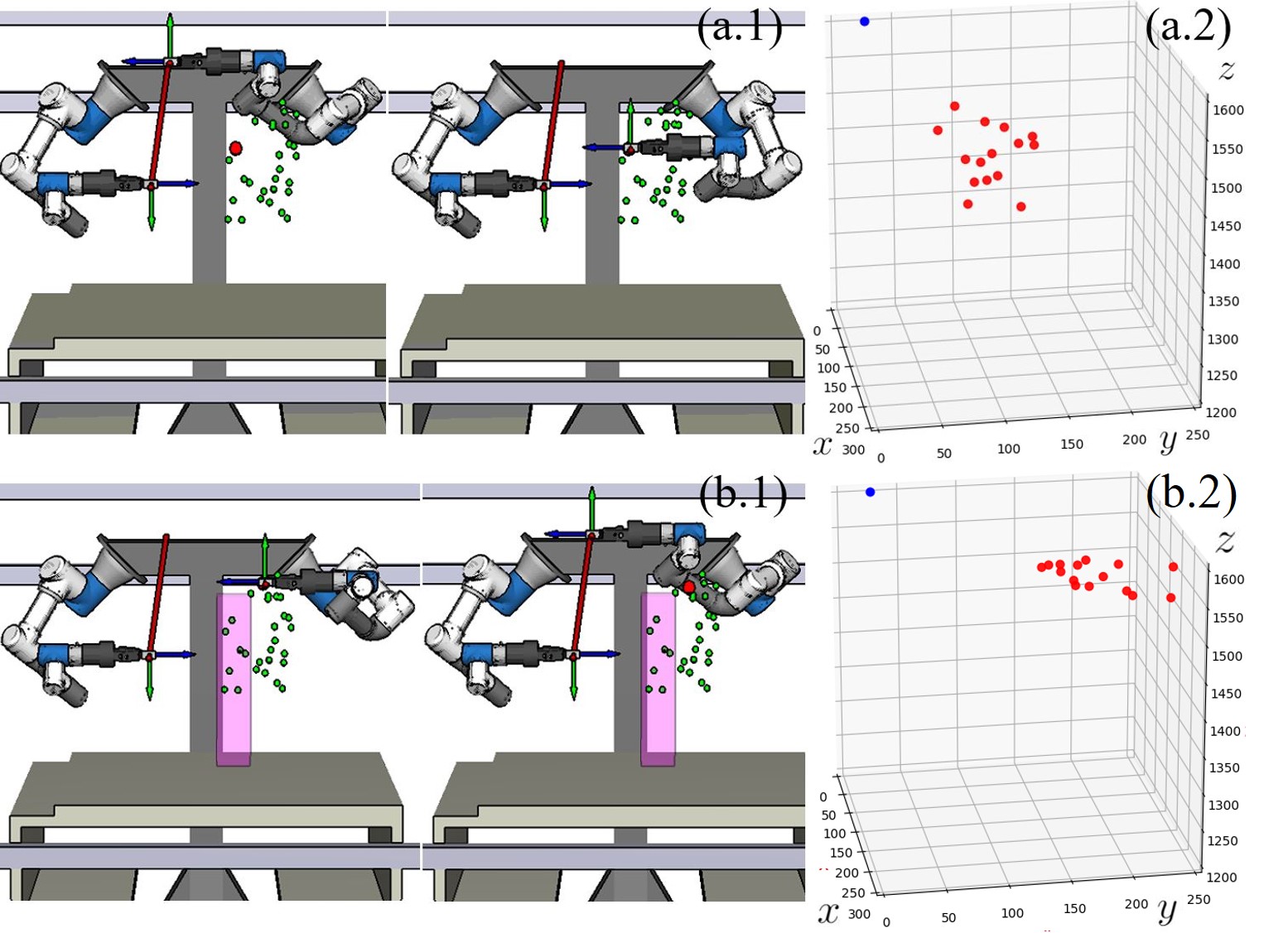}
    \caption{Results when $\omega$ = $[0,0,1]^T$: (a.1) An example of a selected init-goal pose pair. (a.2) Distribution of the selected goals under 15 after times of simulation. (b.1-2) The exemplary selected pair and sample distributions when an obstacle (the purple block) is intentionally placed in the robot workspace.}
    \label{fgrasp}
\end{figure}

In this case, only $f_{grasps}$ affects the rope-pulling motion. This parameter is calculated based on the number of grasping poses simultaneously available to the initial and goal points of a pulling motion. Since the two arms have higher manipulability near the body center, the robot has more feasible grasp poses when a goal point is near the central line. The expectation is validated by Fig.\ref{fgrasp}. In the upper two examples of this figure, the left and right hands pull the rope to points near the center. In the lower two examples, we intentionally included an obstacle at the center for comprehension. When there are obstacles, the number of grasp poses near the center is reduced. As a result, points away from the center are selected as goals. The $f_{grasps}$ parameter allows the robot to select goals considering both kinematic and geometric constraints.

\subsection{Influence of the Constraints and Parameters for Tumbling Optimization}

Second, we study the influence of different constraints and parameter settings on the optimized tumbling motion. We are especially interested in: (1) Necessity of the trajectory constraints \eqref{opsspeed}, \eqref{oppushdir}; (2) Different values of $\boldsymbol{v}_{max}$ and $\gamma$; (3) Different friction coefficients, center of mass (com) positions, and hooking positions. For each of the different constraints and values, we run our algorithms using plates of different sizes (40$mm\times$300$mm$, 150$mm\times$300$mm$, 44$mm\times$500$mm$) and compare the optimized results. The magnitudes of the sizes are coherent with the ones used in the real-world experiments (Table \ref{tab:objects}).

\subsubsection{Necessity of the trajectory constraints \eqref{opsspeed} and \eqref{oppushdir}}
\label{pushconsexperiment}
\begin{figure}[!htbp]
    \centering
    \includegraphics[width=\linewidth]{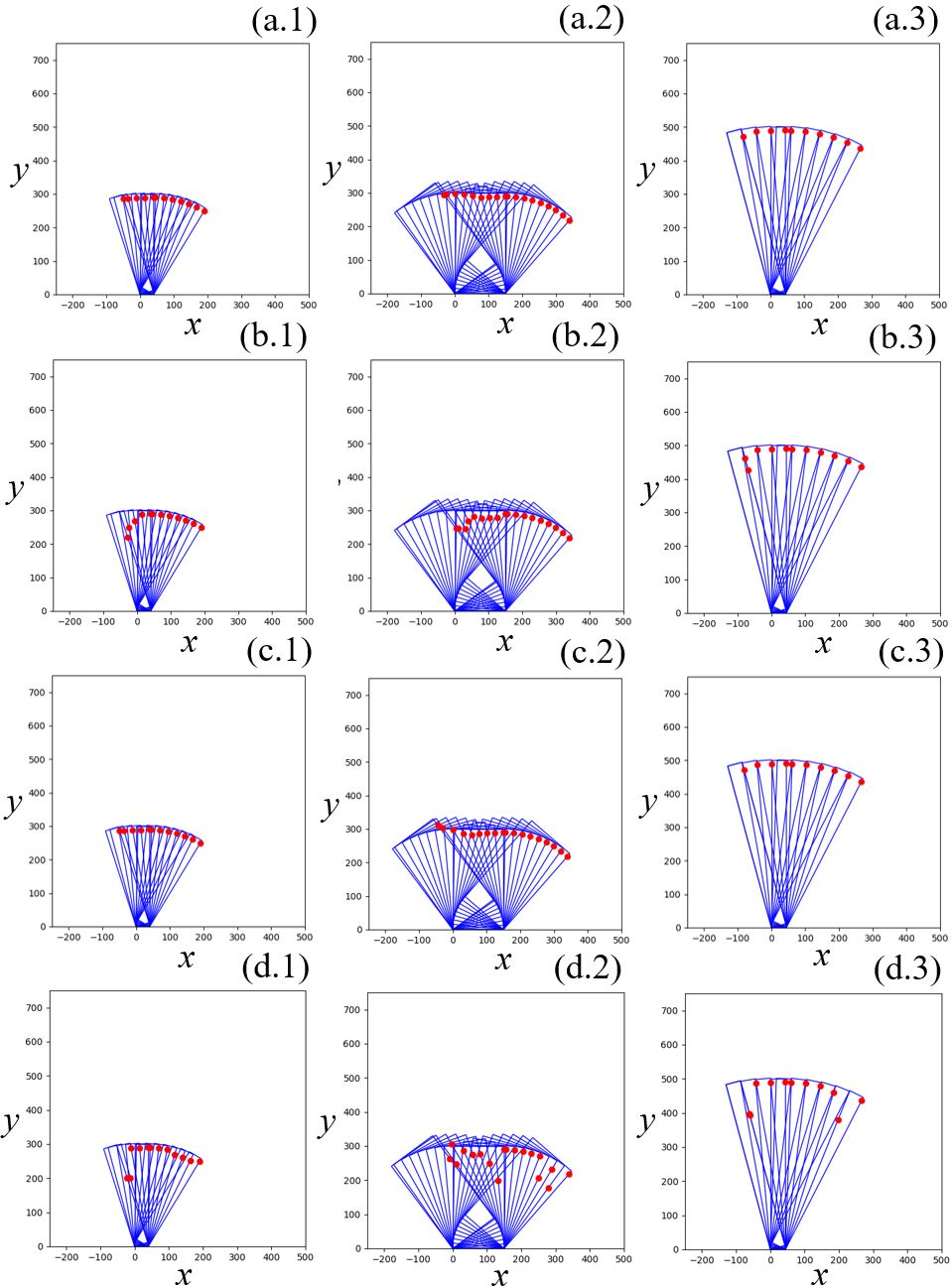}
    \caption{Necessity of the trajectory constraints \eqref{opsspeed} and \eqref{oppushdir}. Each column of the figure uses a plate with a different size (40$mm\times$300$mm$, 150$mm\times$300$mm$, 44$mm\times$500$mm$). The four rows are respectively: (a.1-a.3) Both \eqref{opsspeed} and \eqref{oppushdir} are considered; (b.1-b.3) Only \eqref{opsspeed} is considered; (c.1-c.3) Only \eqref{oppushdir} is considered; (d.4-d.3) Neither \eqref{opsspeed} nor \eqref{oppushdir} is considered.}
    \label{fig:pushpathresult1}
\end{figure}

We study the change of the trajectories under constraints \eqref{opsspeed} and \eqref{oppushdir} and examine their necessity. The results are shown in Fig.\ref{fig:pushpathresult1}. Each column of the figure uses a plate of a different size. The four rows are respectively: (a.1-a.3) Both \eqref{opsspeed} and \eqref{oppushdir} are considered; (b.1-b.3) Only \eqref{opsspeed} is considered; (c.1-c.3) Only \eqref{oppushdir} is considered; (d.4-d.3) Neither \eqref{opsspeed} nor \eqref{oppushdir} is considered. The results imply that both \eqref{opsspeed} and \eqref{oppushdir} are necessary for obtaining a smooth and stable pushing trajectory. When constraint \eqref{oppushdir} is removed, the trajectory gets oscillated. The oscillation can be observed by comparing the trajectories in sub-figure (b.2) with that of (a.2). When constraint \eqref{opsspeed} is removed, the distance between adjacent pushing points becomes large. The large distance can be easily observed by comparing the second half of the trajectory in (c.2) with that of (a.2). When both \eqref{opsspeed} and \eqref{oppushdir} are omitted, the optimized trajectories involve jitters and jerks and become unsuitable for robotic execution. Note that in getting these results, we set the parameters $\boldsymbol{v}_{max}$ = $30mm$/$s$, and $\gamma$ = $20^{\circ}$ when they are used.

\subsubsection{Different values of $\boldsymbol{v}_{max}$ and $\gamma$}
\begin{figure}[!htbp]
    \centering
    \includegraphics[width=\linewidth]{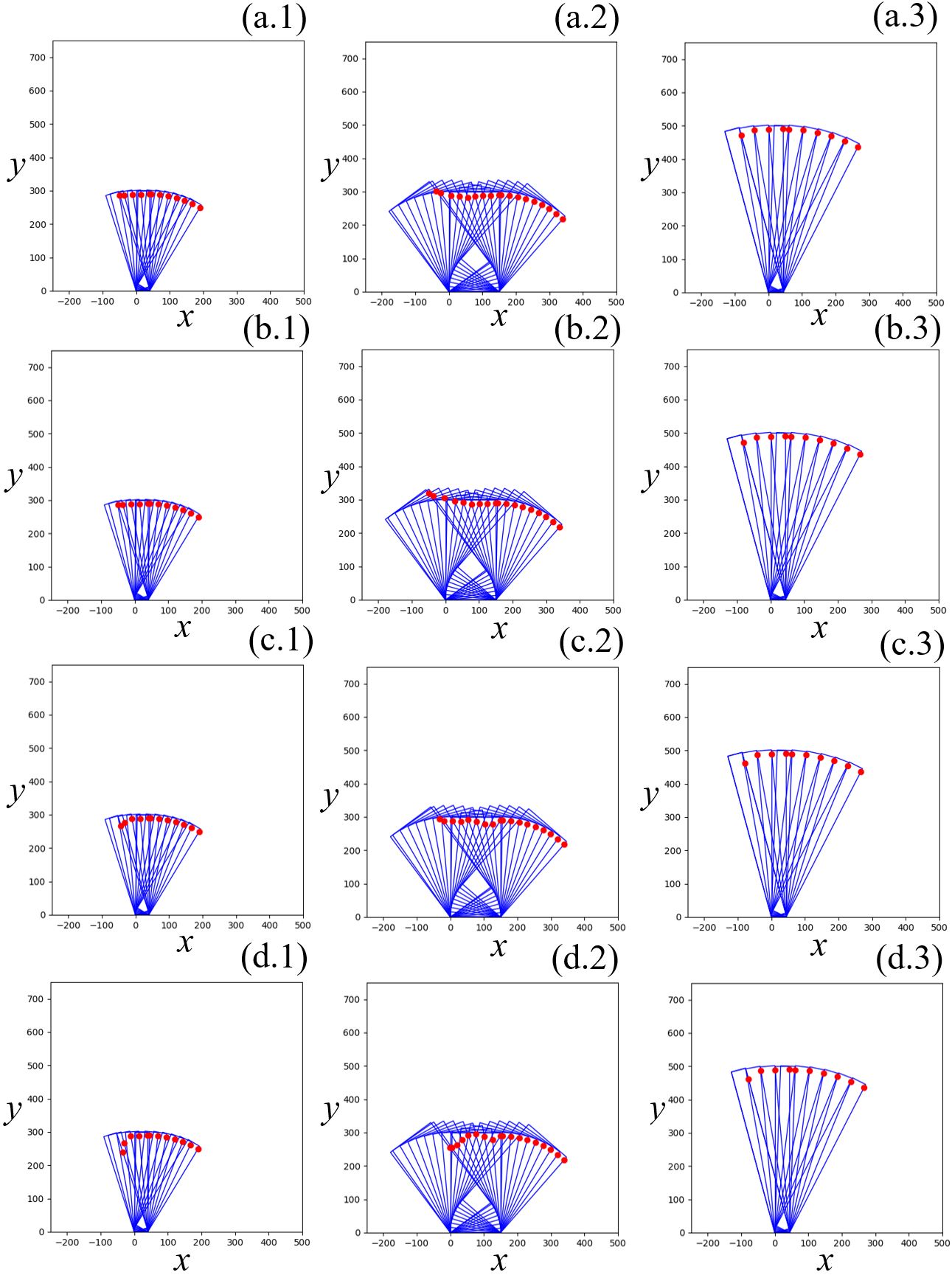}
    \caption{Influence of $\boldsymbol{v}_{max}$ and $\gamma$ values on the optimized trajectories. (a.1-a.3) and (b.1-b.3): The $\gamma$ parameter in constraint \eqref{oppushdir} is changed, while $\boldsymbol{v}_{max}$ is fixed; (c.1-c.3) and (d.1-d.3): The value of $\boldsymbol{v}_{max}$ in constraint \eqref{opsspeed} is changed and \eqref{oppushdir} is fixed.}
    \label{fig:pushpathresult2}
\end{figure}

In this part, we investigate the influence of the parameters $\boldsymbol{v}_{max}$ and $\gamma$ in the trajectory constraints on the optimization. The resulted trajectories using different $\boldsymbol{v}_{max}$ and $\gamma$ values are shown in Fig.\ref{fig:pushpathresult2}. Like Fig.\ref{fig:pushpathresult1}, each column of the figure uses a plate with a different size. For the results in the upper two rows, the $\gamma$ parameter in constraint \eqref{oppushdir} is changed, while $\boldsymbol{v}_{max}$ is fixed. For the results in the lower two rows, the value of $\boldsymbol{v}_{max}$ in constraint \eqref{opsspeed} is changed, and \eqref{oppushdir} is fixed. 

For Fig.\ref{fig:pushpathresult2}(a.1) and (b.1), we set $\boldsymbol{v}_{max}$ = $45mm/s$ and $60mm/s$ respectively, and fixed $\gamma$ = $20^{\circ}$. No significant change is observed in this case as the constraint \eqref{oppushdir} plays a master role. Likewise, for (a.3) and (b.3), we set $\boldsymbol{v}_{max}$ = $60mm/s$ and $75mm/s$, and fixed $\gamma$ = $20^{\circ}$. There is no significant change observed. For (a.2) and (b.2), we set $\boldsymbol{v}_{max}$ = $50mm/s$ and $70mm/s$, and fixed $\gamma$=$20^{\circ}$. In this case, the distances between adjacent pushing points become larger. The observation shows that the constraint \eqref{opsspeed} is effective.

Specifically, for Fig.\ref{fig:pushpathresult2}(c.1) and (d.1), we set $\gamma$ = $40^{\circ}$ and $60^{\circ}$ respectively, and fixed $\boldsymbol{v}_{max}$ = $30mm/s$. By comparing them, we can observe that as the value of $\gamma$ increases, the trajectory biased downward. For Fig.\ref{fig:pushpathresult2}(c.2) and (d.2), we set $\gamma$ = $30^{\circ}$ and $50^{\circ}$ respectively, and fixed $\boldsymbol{v}_{max}$ = $30mm/s$. By comparing them, we can observe that as the value of $\gamma$ increases, the degree of oscillation became stronger. For Fig.\ref{fig:pushpathresult2}(c.3) and (d.3), we set $\gamma$ = $40^{\circ}$ and $60^{\circ}$ respectively, and fixed $\boldsymbol{v}_{max}$ = $45mm/s$. The results show that the downward bias becomes slightly larger in the second half of the trajectory (although no significant difference is observable). The observation indicates that the constraint \eqref{oppushdir} is effective.

\subsubsection{Different friction coefficients, center of mass (com) positions, and hooking positions}
\begin{figure}[!htbp]
    \centering
    \includegraphics[width=\linewidth]{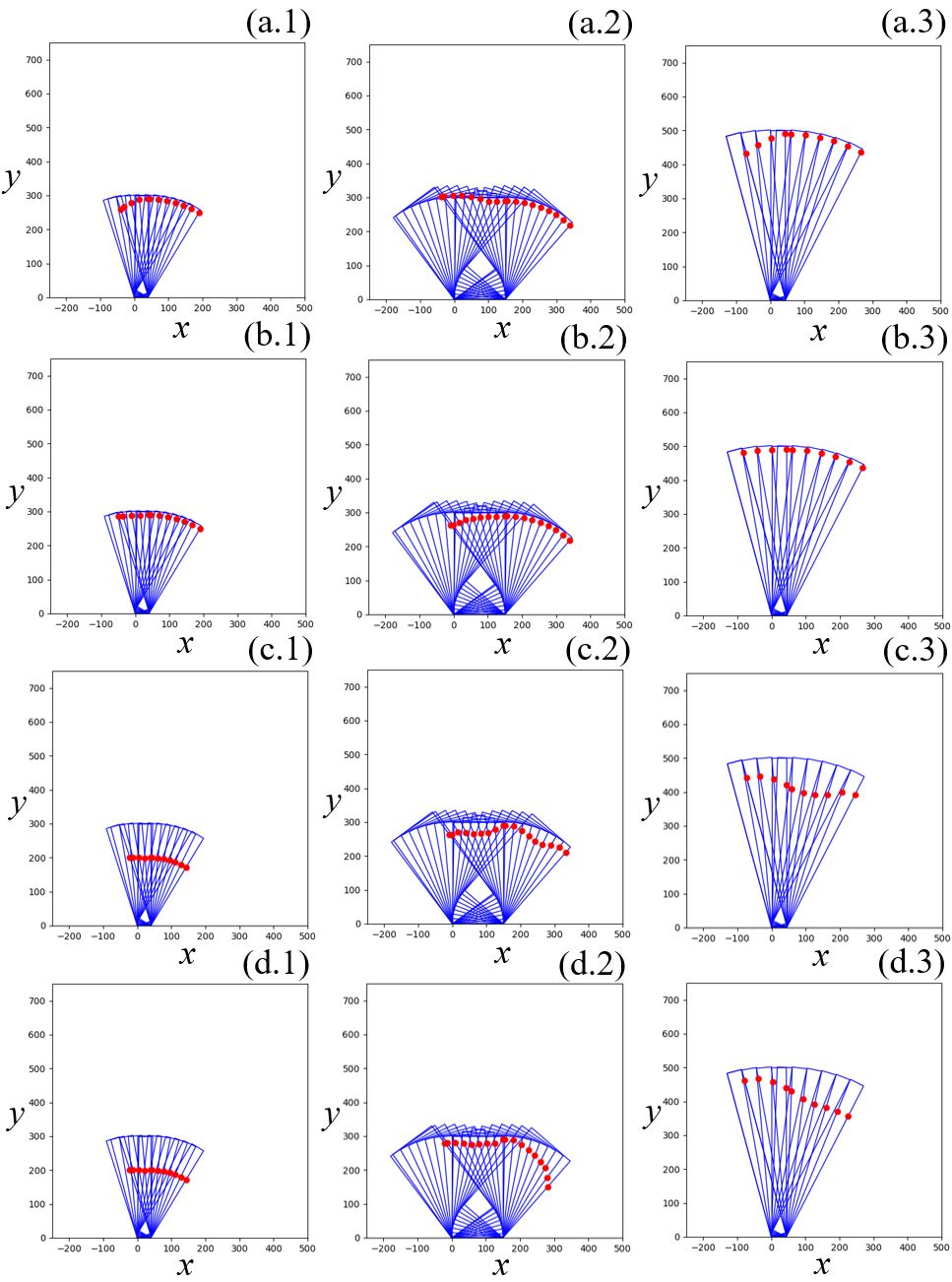}
    \caption{Influence of different friction coefficient ($\mu_1$), center of mass (com) position ($\boldsymbol{r}_{g}$), and hooking position ($\boldsymbol{r}_{h}$) on the optimized trajectories. (a.1-a.3) $\mu_1$ = 0.05; (b.1-b.3) $\mu_1$ = 0.6; (c.1-c.3) $\boldsymbol{r}_{g}$ is changed to the $\boldsymbol{r}_{g_2}$ shown in Fig.\ref{fig:rg_and_rt}; (d.1-d.3) $\boldsymbol{r}_{h}$ is changed to the $\boldsymbol{r}_{h_1}$ shown in Fig.\ref{fig:rg_and_rt}.}
    \label{fig:pushpathresult3}
\end{figure}
In this section, we investigate the changes in trajectory when the friction coefficient at the pushing point ($\mu_1$), center of mass (com) position ($\boldsymbol{r}_{g}$), or hooking position ($\boldsymbol{r}_{h}$) vary. The results are shown in Fig.\ref{fig:pushpathresult3}. Each column of the figure uses a plate of a different size. For Fig.\ref{fig:pushpathresult3}(a.1), (a.2), and (a.3), $\mu_{1}$ is set to 0.05. When the plates are thin, such as (a.1) and (a.3), the optimized trajectories bias downward in the second half. In contrast, when plates are thick, like the one shown in (a.2), the optimized trajectory biases upward. For Figs.\ref{fig:pushpathresult3} (b.1), (b.2), and (b.3), $\mu_{1}$ is changed to 0.6. Compared with the results of 0.05, the optimized trajectories kept neutral in (b.1) and (b.3) but biased downward in (b.2). The results show that $\mu_1$ significantly influences the choices of pushing points in the second half of the trajectory.
\begin{figure}[!htbp]
  \begin{minipage}[c]{0.43\linewidth}
    \includegraphics[width=\textwidth]{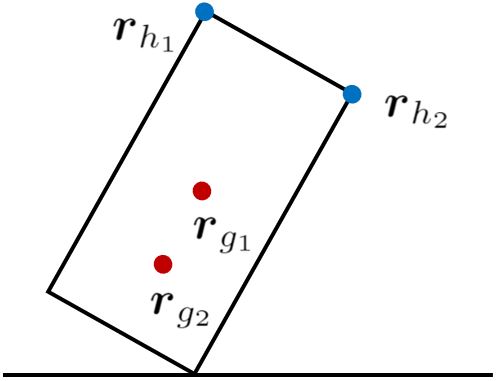}
  \end{minipage}~
  \begin{minipage}[c]{0.54\linewidth}
    \caption{We study the influence of the center of mass (com) position ($\boldsymbol{r}_{g}$) and hooking position ($\boldsymbol{r}_{h}$) on the optimized trajectories by varying each of them to two different positions. These positions are denoted by $\boldsymbol{r}_{g_1}$, $\boldsymbol{r}_{g_2}$, $\boldsymbol{r}_{h_1}$, and $\boldsymbol{r}_{h_2} $in the figure.}
    \label{fig:rg_and_rt}
  \end{minipage}
\end{figure}

Next, we investigate the influences of $\boldsymbol{r}_{g}$ and $\boldsymbol{r}_{h}$. We especially consider two choices for them, as shown by $\boldsymbol{r}_{g_1}$, $\boldsymbol{r}_{g_2}$, $\boldsymbol{r}_{h_1}$, and $\boldsymbol{r}_{h_2}$ in Fig.\ref{fig:rg_and_rt}. $\boldsymbol{r}_{g_1}$ and $\boldsymbol{r}_{g_2}$ are the geometric center and 1/4th of the geometric center of the plate rectangle. $\boldsymbol{r}_{h_1}$ and $\boldsymbol{r}_{h_2}$ are the top-left and top-right corners of the plate rectangle. All previous trajectories were obtained considering $\boldsymbol{r}_{g_1}$ and $\boldsymbol{r}_{h_1}$. In this part, we change them to the other positions for comparison. Fig.\ref{fig:pushpathresult3}(c.1), (c.2), and (c.3) show the trajectories with respect to $\boldsymbol{r}_{g_2}$ and $\boldsymbol{r}_{h_1}$. The results show that when the com is shifted to the bottom of the plate, the maximum duty of a robot hand gets smaller and the hand adjusts itself in wider range. Fig.\ref{fig:pushpathresult3}(d.,1), (d.2), and (d.3) show the trajectories with respect to $\boldsymbol{r}_{g_1}$ and $\boldsymbol{r}_{h_2}$. For all of them, the starting positions get lower. This may be because the robot hand bears a smaller force when pushing starts from a point far from hooking position.

\subsection{Results of Various Plates}
\label{experimentA}
\begin{figure}[!htbp]
    \centering
    \includegraphics[width=.85\linewidth]{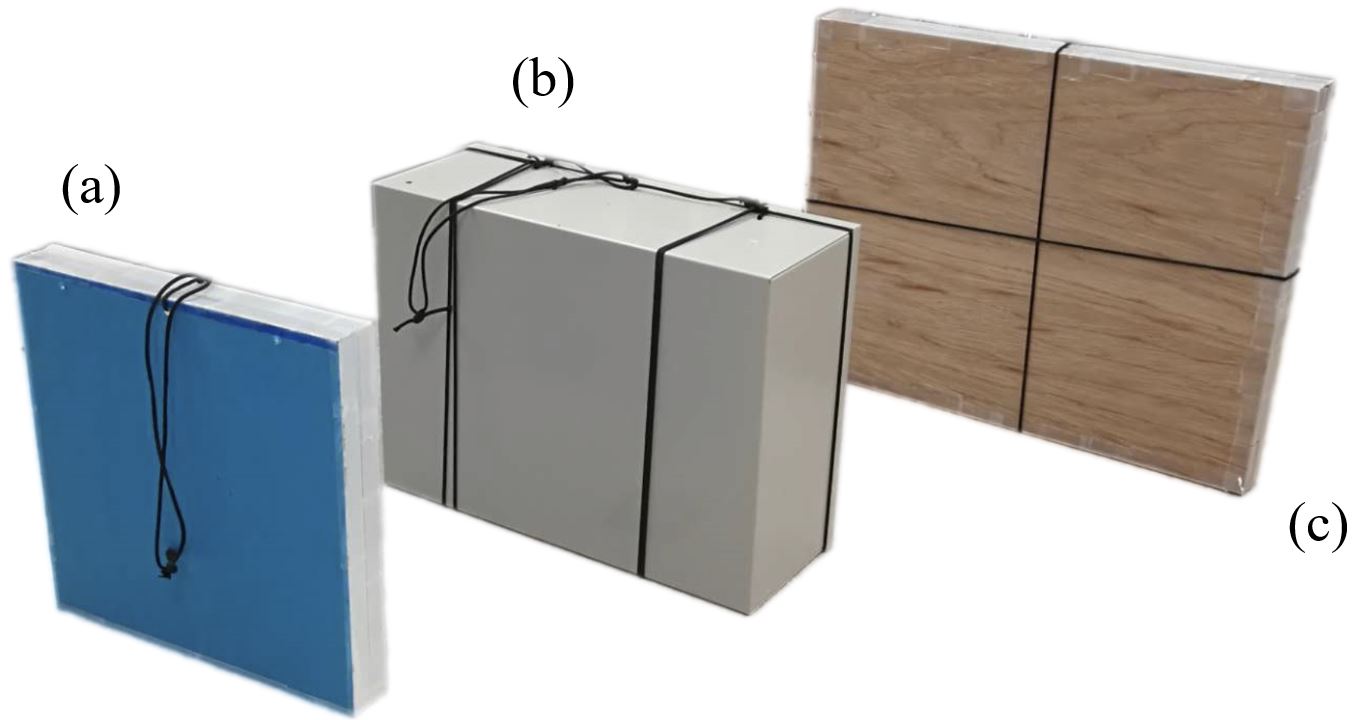}
    \caption{Three different plates used in the experiments. (a) An Acrylic Board. (b) A Stainless Box. (c) A Plywood Plate. Their various parameters are shown in Table \ref{tab:objects}.}
    \label{fig:object}
\end{figure}

Third, we examine the performance of the proposed method using three plates shown in Fig.\ref{fig:object} -- An acrylic board, a stainless box, and a plywood board. The parameters of these plates are shown in Table \ref{tab:objects}. We assume a plate initially lies on the table in front of the robot and the crane hook is pre-connected. 
\begin{table}[!htbp]
    \renewcommand{\arraystretch}{1.2}
    \caption{Parameters of the Plates Used in Experiments}
    \centering
    \begin{threeparttable}
    \begin{tabular}{llllll} \toprule
        \textit{Plate Name} & \textit{h, l, w (mm)} & \textit{m (kg)} & $\boldsymbol{r}_g$ & $\mu_0$ & $\mu_1$ \\ \midrule
         Acrylic Board & 300, 300, 40 & 4.0 & Geom.Cent. & 0.5 & 0.4 \\
         Stainless Box & 300, 400, 150 & 6.0 & Geom.Cent. & 0.4 & 0.1 \\
         Plywood Board & 500, 400, 44 & 6.4 & Geom.Cent. & 0.6 & 0.3 \\ \bottomrule
    \end{tabular} 
    \begin{tablenotes}
    \item[*] $l$ - length; $w$ - width; $h$ - height; $m$ - mass.
    \end{tablenotes}
    \end{threeparttable}
    \label{tab:objects}
\end{table}

\begin{figure*}[!htbp]
    \centering
    \includegraphics[width=\textwidth]{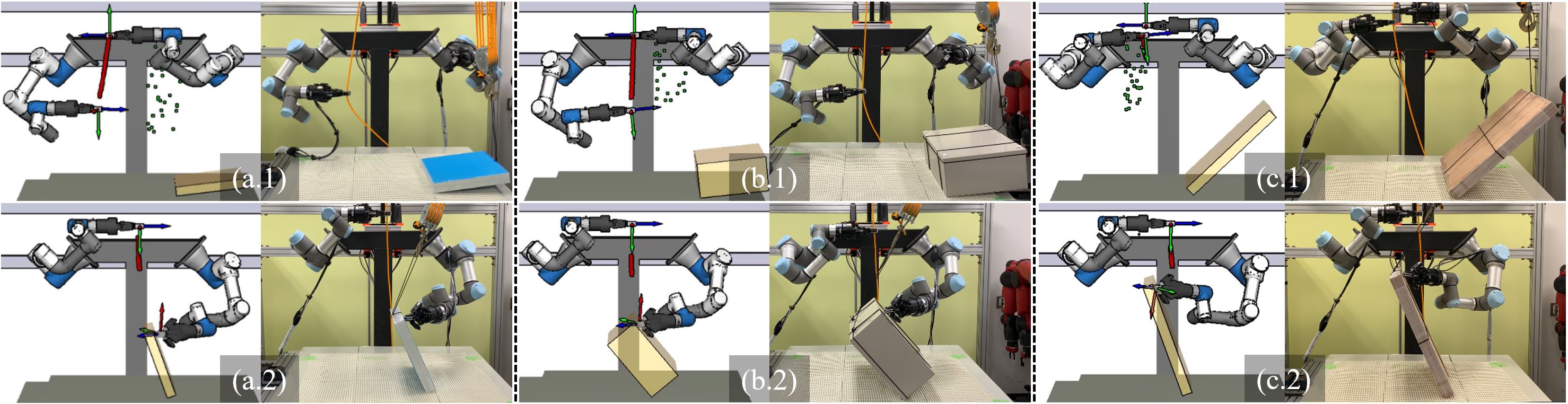}
    \caption{Snapshots of the robotic executions for the three plates ($\omega$ = $[1,1,1]^T$; Watch our supplementary video for details).}
    \label{figexp}
\end{figure*}

Fig.\ref{figexp} shows some snapshots of both the planning environment and real-world execution results for the three plates. In these cases, the weight of the goal quality function is chosen as $\omega$ = $[1,1,1]^T$. The robot managed to pull up the plates with satisfying performance. Readers are encouraged to watch the supplementary video submitted together with this manuscript to better observe the optimized lifting up actions and tumbling trajectories for the different plates.

The most important points we are interested in for the real-world execution are the number of pulling actions, the pulling distances, and the forces born by the pulling hands. We show them in detail in Tables \ref{tabavg}-\ref{tabtime}. 

First, in Table \ref{tabavg}, we compare the execution results of $\omega$ = $[1,1,1]^T$ with other three extreme candidates $\omega$ = $[1,0,0]^T$, $\omega$ = $[0,1,0]^T$, and $\omega$ = $[0,0,1]^T$. The results indicate that when the weight is chosen as $\omega$ = $[1,1,1]^T$, the robot works with a smaller number of actions and relatively large average pulling distances. Meanwhile, the hands bear moderate pulling forces. When the weight is chosen to be $\omega$ = $[1,0,0]^T$, the rope's pulling length becomes dominantly large, and the robot hands bear more forces. When the weight is chosen as $\omega$ = $[0,1,0]^T$, the forces born by the hands are highly suppressed, but the number of actions increase. When the weight is chosen as $\omega$ = $[0,0,1]^T$, there is no significant influence on the length and force values. However, without the parameter $f_{manip}$, the robot and obstacles' collisions cannot be taken into account.

% \begin{table*}[!htbp]
%     \renewcommand{\arraystretch}{1.2}
%     \caption{Comparison of the Real-World Pulling Actions Under Different Weights}
%     \centering
%     \begin{tabular}{@{\extracolsep{3pt}}lllllllllllll} \toprule
%           & \multicolumn{3}{c}{$w$=[1,1,1]$^T$} & \multicolumn{3}{c}{$w$=[1,0,0]$^T$} & \multicolumn{3}{c}{$w$=[0,1,0]$^T$} & \multicolumn{3}{c}{$w$=[0,0,1]$^T$} \\
%          \cmidrule{2-4} \cmidrule{5-7} \cmidrule{8-10} \cmidrule{11-13}
%          \textit{Plate Name} & \textit{\# Actn.} & \textit{Avg. Dist.} & \textit{Forces} & \textit{\# Actn.} & \textit{Avg. Dist.} & \textit{Forces} & \textit{\# Actn.} & \textit{Avg. Dist.} & \textit{Forces} & \textit{\# Actn.} & \textit{Avg. Dist.} & \textit{Forces} \\ \midrule
%          Acrylic Board & L2, R3 & -, - & -, - - & -, - & -, - & -, - & -, - & -, - & -, - & -, - & -, - & -, - \\ 
%          Stainless Box & L4, R3 & -, - & -, - & -, - & -, - & -, - & -, - & -, - & -, - & -, - & -, - & -, -  \\ 
%          Polywood Board & L4, R1 & -, - & -, - & -, - & -, - & -, - & -, - & -, - & -, - & -, - & -, - & -, - \\ \bottomrule
%     \end{tabular}
%     \label{tabavg}
% \end{table*}

\begin{table*}[!htbp]
    \renewcommand{\arraystretch}{1.2}
    \caption{Comparison of the Real-World Pulling Actions Under Different Weights}
    \centering
    \begin{tabular}{@{\extracolsep{3pt}}lllllll} \toprule
           & \multicolumn{3}{c}{$w$~=~[1,1,1]$^T$} & \multicolumn{3}{c}{$w$~=~[1,0,0]$^T$} \\
         \cmidrule{2-4} \cmidrule{5-7}
         \textit{Plate Name} & \textit{\# Actn.} & \textit{Avg.Dist. (L, R) (mm)} & \textit{Forces (L, R) (N)} & \textit{\# Actn.} & \textit{Avg.Dist. (L, R) (mm)} & \textit{Forces (L, R) (N)} \\ \midrule
         Acrylic Board & L2, R3 & (367.68, 430.06) & (3.98, 6.42) & L2, R3 & (448.50, 456.41) & (6.64, 7.48) \\ 
         Stainless Box & L2, R3 & (294.99, 429.37) & (5.70, 6.51) & L2, R3 & (350.41, 422.39) & (5.83, 7.17) \\ 
         Polywood Board & L3, R6 & (332.65, 388.62) & (6.07, 8.12) & L3, R6 & (411.65, 437.13) & (8.21, 8.12) \\
         \midrule
         Total Average & 3.17 & 373.90 & 6.13 & 3.17 & 421.08 & 7.24\\
         \bottomrule
         \\ 
         \toprule
          & \multicolumn{3}{c}{$w$=[0,1,0]$^T$} & \multicolumn{3}{c}{$w$=[0,0,1]$^T$} \\
         \cmidrule{2-4} \cmidrule{5-7}
         \textit{Plate Name} & \textit{\# Actn.} & \textit{Avg.Dist. (L, R) (mm)} & \textit{Forces (L, R) (N)} & \textit{\# Actn.} & \textit{Avg.Dist. (L, R) (mm)} & \textit{Forces (L, R) (N)} \\ \midrule
         Acrylic Board & L3, R3 & (348.71, 340.69) & (4.93, 3.86) & L4, R5 & (283.86, 257.14) & (5.59, 5.96) \\ 
         Stainless Box & L3, R3 & (236.63, 350.60) & (5.64, 5.94) & L4, R4 & (288.68, 297.82) & (7.17, 5.92) \\ 
         Polywood Board & L3, R7 & (334.52, 327.46) & (5.99, 5.86) & L4, R9 & (297.88, 296.20) & (6.62, 7.45) \\
         \midrule
         Total Average & 3.67 & 323.10 & 5.37 & 5.00 & 286.93 & 6.45\\
         \bottomrule
    \end{tabular}
    \label{tabavg}
\end{table*}

Table \ref{tabavg_detail} further shows the details of pulling lengths and forces for each pulling action when $w$ is chosen as [$1$,$1$,$1$]$^T$. The robot alternatively used its left and right arms to pull up the acrylic board and the stainless box. For the plywood board, the robot alternatively used the two arms in the beginning but switched to right-alone actions after step 6. The reason is the plywood board is a large plate. It blocked the right arm action after being to lifted to a high pose.
\begin{table*}[!htbp]
    \renewcommand{\arraystretch}{1.2}
    \caption{Detailed Pulling Length and Forces of Each Pulling Action Using $w$=[$1$,$1$,$1$]$^T$}
    \centering
    \begin{tabular}{@{\extracolsep{6pt}}lllllllllll} \toprule
         \textit{Plate Name} & \textit{Item} & 1 & 2 & 3 & 4 & 5 & 6 & 7 & 8 & 9 \\ \midrule
         Acrylic Board & Actn. Arm & Rgt & Lft & Rgt & Lft & Rgt & - & - & - & - \\
          & Distance ($mm$) & 443.64 & 377.75 & 471.32 & 357.61 & 375.23 & - & - & - & - \\ 
          & Force ($N$) & 8.48 & 4.3 & 5.58 & 3.65 & 5.72 & - & - & - & - \\ 
          & Time ($s$) & 7.57 & 6.09 & 4.02 & 5.94 & 4.36 & - & - & - & - \\ \midrule
        Stainless Box & Actn. Arm & Rgt & Lft & Rgt & Lft & Rgt & - & - & - & - \\
          & Distance ($mm$) & 449.99 & 328.19 & 475.72 & 261.78 & 362.41 & - & - & - & - \\ 
          & Force ($N$) & 7.80 & 5.45 & 5.71 & 5.95 & 6.01 & - & - & - & - \\ 
          & Time ($s$) & 6.41 & 8.45 & 4.08 & 5.87 & 3.96 & - & - & - & - \\\midrule
        Polywood Board & Actn. Arm & Rgt & Lft & Rgt & Lft & Rgt & Lft & Rgt & Rgt & Rgt \\
          & Distance ($mm$) & 455.69 & 367.77 & 449.4 & 343.95 & 425.74 & 286.24 & 394.68 & 394.94 & 211.28 \\ 
          & Force ($N$) & 10.79 & 6.08 & 8.6 & 5.88 & 5.49 & 6.26 & 8.21 & 7.22 & 9.41 \\ 
          & Time ($s$) & 8.55 & 5.39 & 4.00 & 5.99 & 4.47 & 6.36 & 3.99 & 4.15 & 3.85 \\ \bottomrule
    \end{tabular}
    \label{tabavg_detail}
\end{table*}

Table \ref{tabtime} shows the detailed average time costs for the executions of lifting and tumbling each plate using $\omega$ = $[1,1,1]^T$. The most time-consuming part of the proposed planner is optimizing the init-goal pair, including sampling goal points, evaluating goal grasps, and reasoning shared initial and goal grasp poses. They have to be performed on all sampled points to get the defined qualities. The QP solver and RRT planner used to generate the movement and pulling actions also take some time to search and optimize.
\begin{table}[!htbp]
    \renewcommand{\arraystretch}{1.2}
    \caption{Time Costs for Each Part of A Pulling Action}
    \centering
    \begin{tabular}{@{\extracolsep{1pt}}l@{\extracolsep{6pt}}llllll} \toprule
          & \multicolumn{2}{c}{Acrylic} &  \multicolumn{2}{c}{Stainless} & \multicolumn{2}{c}{Polywood} \\ 
         \cmidrule{2-3} \cmidrule{4-5} \cmidrule{6-7}
         $Items$ & ~~Rgt & Lft & Rgt & Lft & Rgt & Lft \\ \midrule
         Initial Grasps ($s$) & ~~0.16 & 0.59 & 0.15 &0.58 & 0.18 & 0.55 \\ 
         Init-Goal Pairs ($s$) & ~~3.02 & 4.31 & 3.13 & 4.28 & 3.13 & 4.38 \\ 
         QP + RRT ($s$) & ~~1.41 & 1.76 & 1.43 & 1.78 & 1.48 & 1.69 \\ \bottomrule
    \end{tabular}
    \label{tabtime}
\end{table}

% The trajectories of the tumbling trajectories for the three plates are shown in Fig.\ref{}. 

%%%%%%%%%%%%%%%%%%%%%%%%%%%%%%%%%%%%%%%%%%%%%%%%%%%%%%%%%%%%%%5
\section{Conclusions}
\label{section6}
This paper proposed an optimization and planning method for a dual-arm robot to lift up and tumble heavy plates using crane pulley blocks. The optimal pulling actions for the pulley bocks were achieved by maximizing or minimizing the length of a pulling motion, the load born by the pulling hand, and the chances of finding a success motion. Visual detection was included in each optimization and execution loop to update the states of the plate. The optimal tumbling was implemented by considering the force and moment applied to the plate. An optimal sliding-push trajectory was planned by minimizing the forces needed to maintain equilibrium. After tumbling, the plate was lowered down onto the table to completely finish a task. The action sequences and motion details of both pulling and tumbling were autonomously decided following respective optimizations. Experiments and analysis showed that the optimizations responded well to changing plate sizes, weights, materials, etc. The robot was able to flexibly and efficiently adapt its action sequences and motion details to different scenarios.

The focus of this manuscript is the optimization and planning aspect. We did not consider the rope and assumed it to be pre-attached to the plates. Autonomously manipulating and attaching a crane rope using the robots and peripherals will be our future work.

% %%%%%%%%%%%%%%%%%%%%%%%%%%%%%%%%%%%%%%%%%%%%%%%%%%%%%%%%%%%
% \section*{Appendix}
% This appendix section presents the numerical implementation for the optimization equations (\ref{opgoal}-\ref{opsspeed}). In implementing the optimization, we consider two numerical discretizations. The first one is the time instant, which has been defined using $t_i$ in Section V. The second one is the contact $\boldsymbol{p}_1$. The contact point $\boldsymbol{p}_1$ at different $t_i$ is modeled using a series of discrete points $\boldsymbol{p}_{1_j}(t_i)$, $j$=$0$,$1$,$2$,$\ldots$. Under this modeling, the optimization problem is formulated as a linear programming problem shown in Fig.\ref{figapp}. The optimization is realized by implementing the linear programming.
% \begin{figure}[!htbp]
%     \centering
%     \includegraphics[width=.7\linewidth]{imgs/appendix.jpg}
%     \caption{The contact point $\boldsymbol{p}_1$ at different $t_i$ is modeled using a series of discrete points $\boldsymbol{p}_{1_j}(t_i)$. Under this modeling, the optimization problem is formulated as a linear programming problem that finds a best combination of the contact points across all $t_i$.}
%     \label{figapp}
% \end{figure}
% The robot kinematic constraint is considered lazily after a trajectory is found by linear programming. Each contact point along the trajectory will be examined by an IK solver. If a point does not have a feasible IK solution, it will be invalidated and the next best trajectory will be tested. The numerical implementation will finally find a optimal trajectory or report a failure. 

\bibliographystyle{IEEEtran}
\bibliography{citations.bib}

\end{document}